\documentclass[preprint,12pt]{elsarticle}

\usepackage{amssymb}

\graphicspath{{./}}
\DeclareGraphicsExtensions{.pdf,.jpeg,.png,.jpg,.tiff}

\usepackage{amsmath}
\usepackage{graphicx}
\usepackage{hyperref}
\usepackage{multirow}
\usepackage{siunitx}
\usepackage{subcaption}
\usepackage{tikz}
\usepackage{picins}

\newdimen\imageheight

\journal{Journal of Computational Science}

\begin{document}

\begin{frontmatter}



\title{Model calibration using a parallel differential evolution algorithm in computational neuroscience: simulation of stretch induced nerve deficit}


\author[datsi,ccs]{Antonio LaTorre\corref{cor1}}
\author[oxfordeng]{Man Ting Kwong}
\author[oxfordmat]{Juli{\'a}n A. Garc{\'i}a-Grajales}
\author[purdue]{Riyi Shi}
\author[oxfordeng]{Antoine J{\'e}rusalem\corref{cor1}}
\author[oxfordeng,lurtis]{Jos{\'e}-Mar{\'i}a Pe{\~n}a\corref{cor1}}

\address[datsi]{DATSI, ETSIINF, Universidad Polit{\'e}cnica de Madrid}
\address[ccs]{Center for Computational Simulation, Universidad Polit{\'e}cnica de Madrid}
\address[oxfordmat]{Mathematical Institute, University of Oxford}
\address[oxfordeng]{Department of Engineering Sciences, University of Oxford}
\address[purdue]{Department of Basic Medical Sciences, College of Veterinary Medicine, Weldon School of Biomedical Engineering, Purdue University}
\address[lurtis]{Lurtis Ltd., Oxford}

\cortext[cor1]{Corresponding authors: a.latorre@upm.es (Antonio LaTorre), jm.penya@lurtis.com (Jos{\'e}-Mar{\'i}a Pe{\~n}a), antoine.jerusalem@eng.ox.ac.uk (Antoine J{\'e}rusalem)}

\begin{abstract}
Neuronal damage, in the form of both brain and spinal cord injuries, is one of the major causes of disability and death in young adults worldwide. One way to assess the direct damage occurring after a mechanical
insult is the simulation of the neuronal cells functional deficits following the mechanical event. In this study, we use a coupled mechanical electrophysiological model with several free parameters that are required to be calibrated against experimental results. The calibration is carried out by means of an evolutionary algorithm (differential evolution, DE) that needs to
evaluate each configuration of parameters on six different damage cases, each of them taking several minutes to compute. To minimise the simulation time of the parameter tuning for the DE, the stretch of one unique fixed-diameter axon with a simplified triggering process is used to speed up the calculations.
The model is then leveraged for the parameter optimization of the more realistic bundle of independent axons, an impractical configuration to run on a single processor computer. To this end, we have developed a parallel implementation based on OpenMP that runs on a multi-processor taking advantage of all the available computational power. The parallel DE algorithm obtains good results, outperforming the best effort achieved by published manual calibration, in a fraction of the time. While not being able to fully capture the experimental results, the resulting nerve model provides a complex averaging framework for nerve damage simulation able to simulate gradual axonal functional alteration in a bundle.
\end{abstract}


\end{frontmatter}





\section{Introduction}
Neuronal damage caused by mechanical insults is the focus of a wide
range of disciplines, ranging from theoretical computational neuroscience
to the study of their secondary effects in the biochemical balance of the brain,
their resulting mid- to long-term cognitive and motor deficits, to their potential treatment. These
insults may vary in intensity and frequency (e.g., explosion blasts, traffic
accidents, or repetitive sports related impacts) as well as in their location (e.g., traumatic brain injury vs. spinal cord injury).
Additionally, the cascade of events resulting from the mechanical damage has also a
timeline of effects that include different forms of tissue degeneration and
cell death, e.g., neuronal calcium homeostasis, metabolism alterations
or physiological dysfunction \cite{Verweij1997,Dumont2001,Park2004}. This paper focuses on the short-term effects
caused by the mechanical damage through the simulation of the mechanically induced
damage of the electrophysiology of the signal transmission in axons. To this end, we
consider here the calibration of an individual axonal transmission and then extend it
to the analysis of the simulation of a whole nerve.

The complexity of the coupling between mechanics and electrophysiology
requires the identification of many parameters. Some of these parameters may be
obtained by controlled experimental studies as reported in the literature, but many
other parameters are not readily available in such tests.
Nonetheless, some experimental studies offer a valuable set of controlled cases
in which meso- or macroscale effects can be measured \cite{Morrison1998,Morrison2006,Shi2006}. The challenge is to identify
those unknown values for the corresponding free parameters that (after simulation)
show the results that best approximate the reference experimental studies.

To accomplish this objective, and given the impossibility of exhaustively exploring
all the possible interactions among these parameters, we propose the use of a
parallel implementation of the differential evolution (DE) algorithm \cite{Storn1997}, a
metaheuristic algorithm that has been previously successfully used to solve complex
scientific and engineering problems \cite{karabouga2004simple,babu2007differential,bjorck2007genx,qing2009differential,tsai2015improved,chin2016accurate}.
This parallel implementation makes use of the
OpenMP API to speed up the evaluation of each candidate solution, as this is
the most time-comsuming part of the algorithms due to the nature of the fitness
function, which requires to run a complete simulation of the electrophysiological model.
Alternatively, GPUs could also be used to parallelize the algorithm, as they
have already exhibited excellent performance for this kind of problems
\cite{Guo2014,Li2015,Yang2015,Li2018}, see Section \ref{sec:conclusions}.

The remainder of this paper is organized as follows. Section \ref{sec:preliminaries}
briefly reviews past related work. In Section \ref{sec:singleaxon}, we present the
simplified single axon problem for the tuning of the optimization framework. We provide a formulation of the problem in
Section \ref{sec:problem}, a description of the experimental scenario under study and
the parameter tuning carried out in Section \ref{sec:scenario} and the results
and most relevant findings for this first simulation experiment in Section \ref{sec:res}. Then,
the more realistic multi-axon calibration problem is presented in Section \ref{sec:axon2nerve}.
Finally, Section \ref{sec:conclusions} concludes this work.

\section{Preliminaries}
\label{sec:preliminaries}
The effects of the traumatic damage in either the brain (traumatic brain injury, TBI) or the spinal cord (spinal cord injury, SCI) are actually the product of a sequence of events \cite{Kuzhandaivel2011}. Firstly, the direct damage produced by the mechanical insult disrupts cells in the region where the mechanical load is applied; secondly, complex regulatory mechanisms are activated that may also cause the degradation of the affected area or even nearby tissue. This potentially leads to cell impairment at the functional level, or even cell death \cite{Norenberg2004,Amar2007}. The way the different effects triggered in this second phase interact and contribute to the degradation process of both TBI and SCI is still a matter of debate. Among these secondary pathophysiology consequences, the most relevant (but not the only ones) are \cite{Dumont2001}: 
\begin{itemize}
\item The massive release of glutamate producing excitotoxicity that induces excessive activation of glutamate receptors, leading to further neuronal cell death \cite{Xu1998,Park2004,Hinzman2016},
\item Ischemia/hypoxia caused by metabolic pathway activated by mitochondrial  dysfunction and apoptosis \cite{Verweij1997,Xiong1999,Dumont2001,Hiebert2015},
\item Axonal demyelination and degeneration \cite{Schwab1996,Shi2015} as a consequence of apoptosis \cite{Abe1999,Li1999},
\item Loss of vascular tone autoregulation (relevant for second-impact syndrome) as well as other vascular changes (e.g., vasospasm and thrombosis) that may produce edemas and other secondary complications \cite{Tator1990,Bareyre2003},
\item Nonketotic metabolic acidosis causing an increase in the incidence of infection and activation of acid-sensing ion channels \cite{Hu2011,Wemmie2013},
\item Increased concentration of nitric oxide and free oxygen radicals causing the oxidative death of neurons \cite{Toborek1999}, lipid peroxidation \cite{Chan1984,AbdulMuneer2015} and cytoskeletal degradation \cite{Banik1982}.
\end{itemize}
The injury consequences vary according to the cell type. For instance, neurons are more vulnerable to excitotoxicity derived from parthanatos, but glial cells are more affected by metabolic perturbations producing ischemic/hypoxic apoptosis \cite{Kuzhandaivel2011}.

\subsection{Experimental study of stretch induced axonal functional damage}
\begin{figure}[thpb]
  \centering
  \includegraphics[width=10cm]{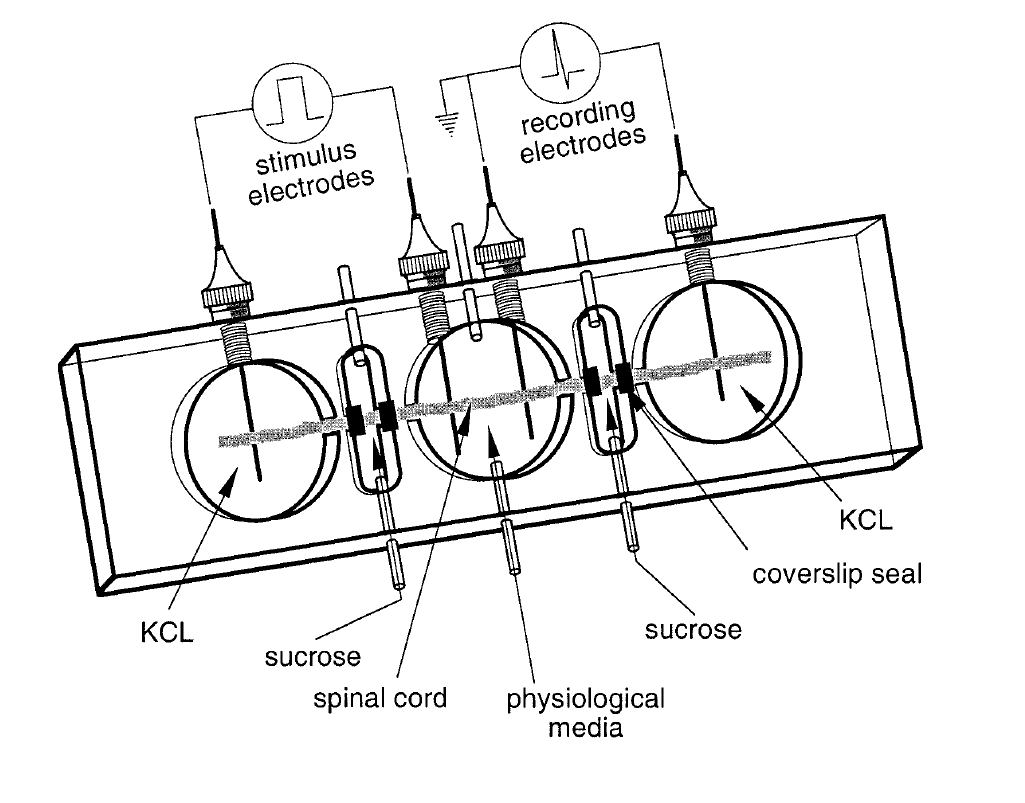}
  \caption[Double sucrose gap experimental set-up.]{Double sucrose gap set-up \cite{Shi1999}.}
  \label{fig:doubleSucrose}
\end{figure}

The experimental studies by Shi et al. \cite{Shi2006} set out to  measure the changes in compound action potential (CAP) amplitude before and after different levels and rates of traumatising stretch of guinea pig spinal cord white matter. The pre-stretch CAP was measured in a double sucrose gap, an experimental rig with three chambers. The ends of the spinal cord segment were placed in two end-chambers, submerged in intracellular ionic solution, while the rest of the spinal cord segment lays in the middle chamber submerged in an extracellular solution. The chambers were separated by concentrated sucrose, acting as a high impedance partition. One of the end chambers had a stimulating electrode to induce action potentials (AP) in the spinal cord segment. The resulting signal can then be measured as a potential difference between the middle and the other end chamber with electrodes, see Figure \ref{fig:doubleSucrose}. In addition to this set-up, an impacting rod was used in the middle chamber to impose different levels and rates of traumatising stretches, while the CAP was then measured continuously after stretch for up to 30 min.

\begin{figure}[thpb]
  \centering
  \includegraphics[width=10cm]{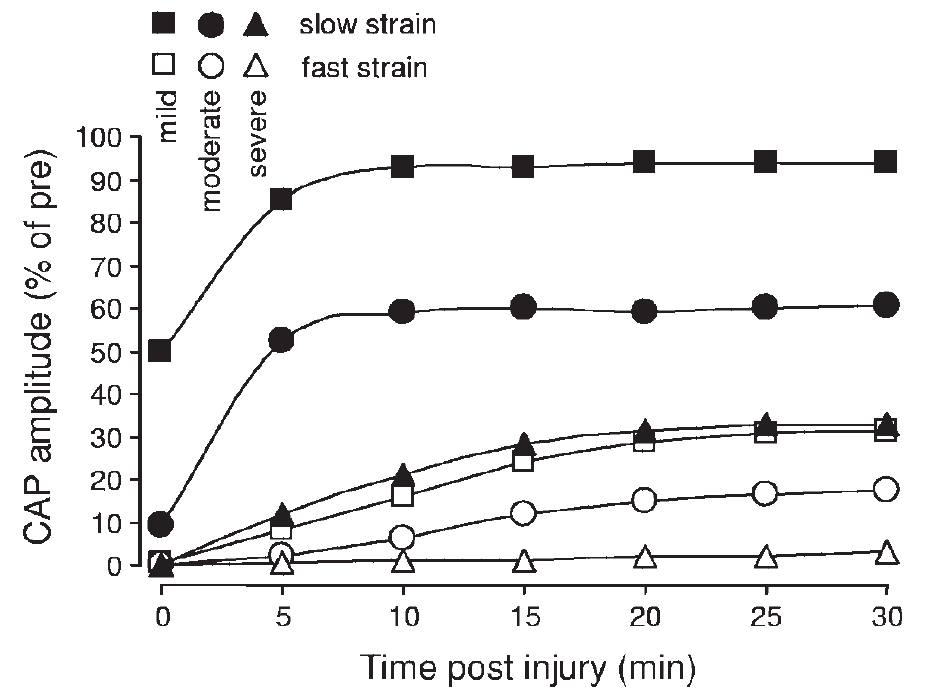}
  \caption[CAP amplitudes of six loading cases during mechanical stress free relaxation]{CAP amplitudes of six loading cases (presented as a percentage of the pre-strain signal) during a \SI{30}{\minute} post-stretch period \cite{Shi2006}.}
  \label{fig:expData}
\end{figure}

The post-stretch CAP as a percentage of the pre-stretch CAP was used as an evaluator of the change in CAP after mechanical stretch. During the experiment, subsequent CAP measurements were also done after the initial injury to observe a recovery of CAP amplitude accompanying a \SI{30}{\minute} stress free relaxation of the spinal cord segment. CAP measurements were measured at \SI{5}{\minute} intervals. In Figure \ref{fig:expData}, the percentage of CAP post-stretch compared to pre-stretch vs. relaxation time is presented. The findings show that a higher strain, and a higher strain rate in general, cause a larger CAP reduction, which is recoverable to varying degrees over \SI{30}{\minute}.

This work is used throughout the paper as the experimental reference.

\subsection{Electrophysiological simulation}
Computational simulations have become an essential tool in the understanding of the most complex functions of the brain. Neuroscience has been using simulations in the study of the electrophysiological aspects of impulse transmission at many different levels from the simulation of networks of spiking neurons \cite{Brette2007} mainly based on single compartment (with models such as NEST \cite{Gewaltig2007}, SpikeNET \cite{Delorme1999}, or NCS \cite{Markram1997}), to multicompartmental models based on the Hodgkin-Huxley model \cite{Hodgkin1952} or Cable Theory \cite{Tuckwell1988} (such as NEURON \cite{Hines1997} or SPLIT \cite{Hammarlund1997}).

More detailed lower scale frameworks have also been proposed to simulate neurotransmitter release \cite{Montes2015} by use of a Monte Carlo particle interaction model developed on top of MCell software \cite{Stiles2001}, or to provide coarse-grain event-driven or integrate-and-fire simulations (e.g., MVASPIKE \cite{Rochel2003}). Finally, to bridge scales, multilevel simulators, such as GENESIS \cite{Bower1995}, range from biochemical interactions at the synaptic level to the simulation of medium-to-large networks or even system-levels.

\subsection{Biomechanical simulations of neuron networks}
Although electrophysiological simulation platforms for the brain have been developed since the mid-nineties based on theoretical models proposed in the fifties, the mechanics of the brain is a matter of much more recent studies. In particular, its interaction with the functional operation of the neuronal networks is an open research area \cite{Goriely2015}.

From a mechanical perspective, there are multiple scales and scenarios under which simulations can be applied: brain surgery, neurogenesis and brain development, brain tumours, and the full range of traumatic damage (both brain and spinal cord \cite{Goriely2015}). In the context of TBI and SCI, there are also multiple phenomena to model, such as fluid dynamics effects (derived from the blood vessel networks and the cerebrospinal and interstitial fluids, as well as their interdependencies \cite{Tully2010}), mechanically induced electrochemistry alterations (due to, for instance, traumatic damage or cardiovascular accidents, e.g., stroke, that cause tissue swelling and brain edemas related to the cytotoxic propagation of the damage \cite{Dronne2006}), and coupled electro-mechanical effects (that concerns primary functional reaction from both neuron and glial cells to a mechanical insult, which may be modeled both at the cellular \cite{Jerusalem2012} and tissue levels \cite{Moore2009,Jean2014}).

In terms of simulation frameworks, there is also a wide range of approaches that consider finite element calculations as the modeling vehicle, focusing on the mechanical properties of the brain tissue \cite{Miller1997,Hrapko2006} to cell body cytoplasm-membrane interactions \cite{Jerusalem2012}, or whole organ simulations (e.g., to model chronic traumatic encephalopathy \cite{Ghajari2017} or different brain surgery procedures \cite{Mendizabal2015,Fletcher2016}).

Few approaches have combined both mechanical and electrophysiological simulation either at the cell \cite{Jerusalem2014,GarciaGrajales2015,Cinelli2018,Kwong2019} or tissue scale \cite{Garcia-Gonzalez2019}. NEURITE \cite{GarciaGrajales2015}, one of the earliest software dedicated to the cell scale is used here as the underlying simulator of nerve damage.

\subsection{Differential evolution and model calibration}
Simulation model calibration is an open research area that has always been dependent on the characteristics of the simulation mechanisms to calibrate. Numerical approaches use properties of the simulation mechanics to identify optimal parameters but are limited to the analytical study of the derivable and continuity properties of the function to optimize. When such numerical approach cannot be carried out, estimation approaches attempt to find a good (maybe not optimal calibration) by indirect methods. Independent parameter calibration can be performed by means of methods such as sensitive analysis \cite{kim2010systematic,tahmasebi2012optimization}, whereas multiple interdependent parameters require approaches used in experiment design, such as Taguchi's method \cite{yan1993diesel}. When the number and the complexity of the relationships among the parameters grow, a global optimization approach is required. Although one of the most commonly used techniques for the global calibration of simulated models is Monte Carlo approaches \cite{vargas2003efficiency} (through in-biased random sampling the parameter space) the search of the optimal calibration of parameters requires a guided search strategy.

The use of heuristic optimization methods in the calibration of simulation models has been a very active research topic \cite{montagna2009parameter,arsenault2013comparison}. Most of the optimization approaches based on heuristic optimization have used genetic algorithms \cite{kim2001genetic,goicoechea2002wavelength}, but many other heuristic approaches have also been used, such as particle swarm optimization \cite{scheerlinck2009calibration,ruiz2015calibration}, ant colony optimization \cite{calvez2007ant}, scatter-search \cite{liberatore2009location}, or local search methods like simulated annealing or tabu search \cite{lidbe2017comparative}. Recently,  hybrid heuristic optimization approaches have also been used for the calibration of computational simulation models, for instance, multiple offspring sampling \cite{pena2019softfem}, or memetic algorithms \cite{paz2015calibration}.

In this work we have selected DE, which has already been used in the parameter calibration of several simulation models \cite{ezquerro2011calibration,plasencia2014geothermal}. It is an efficient continuous optimization technique that is able to work with a relative small population size.

\section{Optimization parameter identification}
\label{sec:singleaxon}

\subsection{Problem formulation}
\label{sec:problem}
This study is based on published work in which NEURITE,
a computational model coupling mechanics and electrophysiology in SCI
\cite{Jerusalem2014} was proposed, see implementation in Ref. \cite{GarciaGrajales2015}\footnote{NEURITE software can be freely downloaded from \url{https://senselab.med.yale.edu/modeldb/showModel.cshtml?model=168861}}.

\subsubsection{Electrophysiological-mechanical model}
\label{sec:em_model}

\begin{figure}[ht]
\centering
\includegraphics[width=1.0\textwidth]{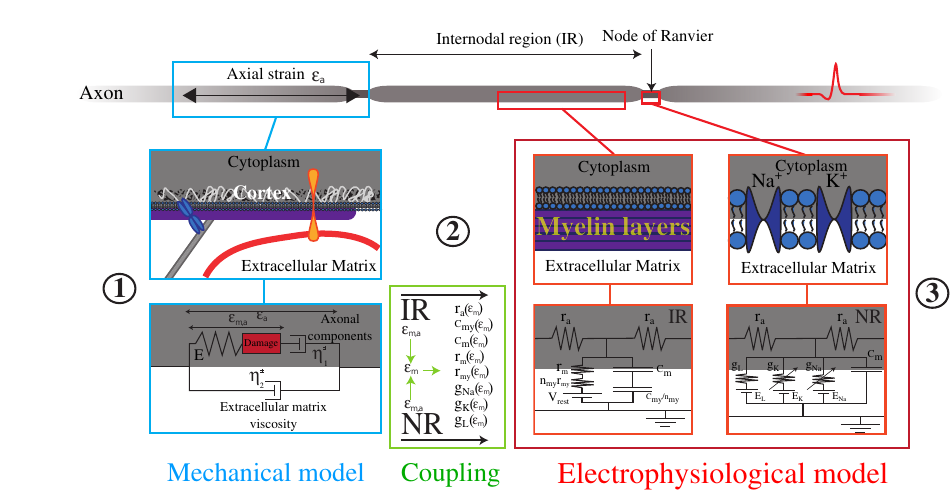}
\caption{The complementary models that simulate the electrophysiological-mechanical
coupling: The mechanical model (1) translates the mechanical loads into geometrical and
functional changes (2) of the parameters used by the electrophysiological model (3) \cite{Jerusalem2014}.}
\label{fig:model}
\end{figure}

The mechanical model described in Ref. \cite{Jerusalem2014} accounts for the
visco-elasto-plastic deformation of the axons embedded in a viscous
extracellular matrix. The electrophysiological model simulates impulse
transmission using Cable Theory for the myelinated internodal regions
(IR) and the Hodgkin-Huxley model for the nodes of Ranvier (NR) in myelinated axons
\cite{Hodgkin1952,Koch1999}.
NEURITE couples both models by a) accounting for the geometrical membrane
alterations ensuing from mechanical deformation, and b) potential damage
in ion channels as observed experimentally through the so-called sodium
ion channel left-shift \cite{Wang2009}, see Ref. \cite{Jerusalem2014} for more
details.
Figure \ref{fig:model} shows both coupled models integrated in the simulation of
the traumatic axonal damage.

Following some assumptions on the electrophysiological parameters, the resulting coupled model was left with six free parameters that need to be adjusted.
The name and description of these parameters is given in Table \ref{table:para_cal},
whereas the interval of possible values for each of these parameters is discussed
in Section \ref{sec:scenario}.

\begin{table}[thpb]
  \begin{center}
  \caption[Model parameters to be calibrated.]{Model parameters to be calibrated \cite{Jerusalem2014}.}
  \label{table:para_cal}
  \begin{tabular}{l l}
    \hline
    Parameter & Definition \\
    \hline
    E & Young's modulus\\
    k & Damage evolution \\
    $\eta_{eq}$ & Equivalent viscous parameter \\
    $\tilde{\varepsilon}$ & Microscopic strain threshold \\
    $\kappa$ & Sensitivity parameter for microscopic relaxation \\
    $\gamma$ & Coupling sensitivity exponent\\
    \hline
  \end{tabular}
  \end{center}
\end{table}

\subsubsection{Application scenario}
The problem is defined as one myelinated axon of arbitrary diameter \SI{3}{\micro\metre}, which is modeled as a series of
one-dimensional elements with alternatively distributed NR and IR segments. The axon
experiences strains of three different magnitudes (mild, moderate
and severe) at two different rates (fast and slow). These six combinations
define the damage cases under study. For each of the damage cases, the simulation calculates
the AP originated from one fixed transmitted signal at the beginning of
the axon at a point further down the axon. These signal transmissions are recorded at seven
different time-points (just after the mechanical insult and 5, 10, 15, 20, 25,
and 30 min after). For a given time-point and a given loading case,
the final output consist of the ratio of the maximum recorded AP amplitude for this
damaged case to the maximum recorded amplitude for the undamaged case (1 means no damage,
0 means completely damaged). These AP ratios are then compared to the CAP from the experiments
(see Ref. \cite{Jerusalem2014} for more details about the assumptions).

\begin{figure}[ht]
  \centering
  \includegraphics[width=0.7\textwidth]{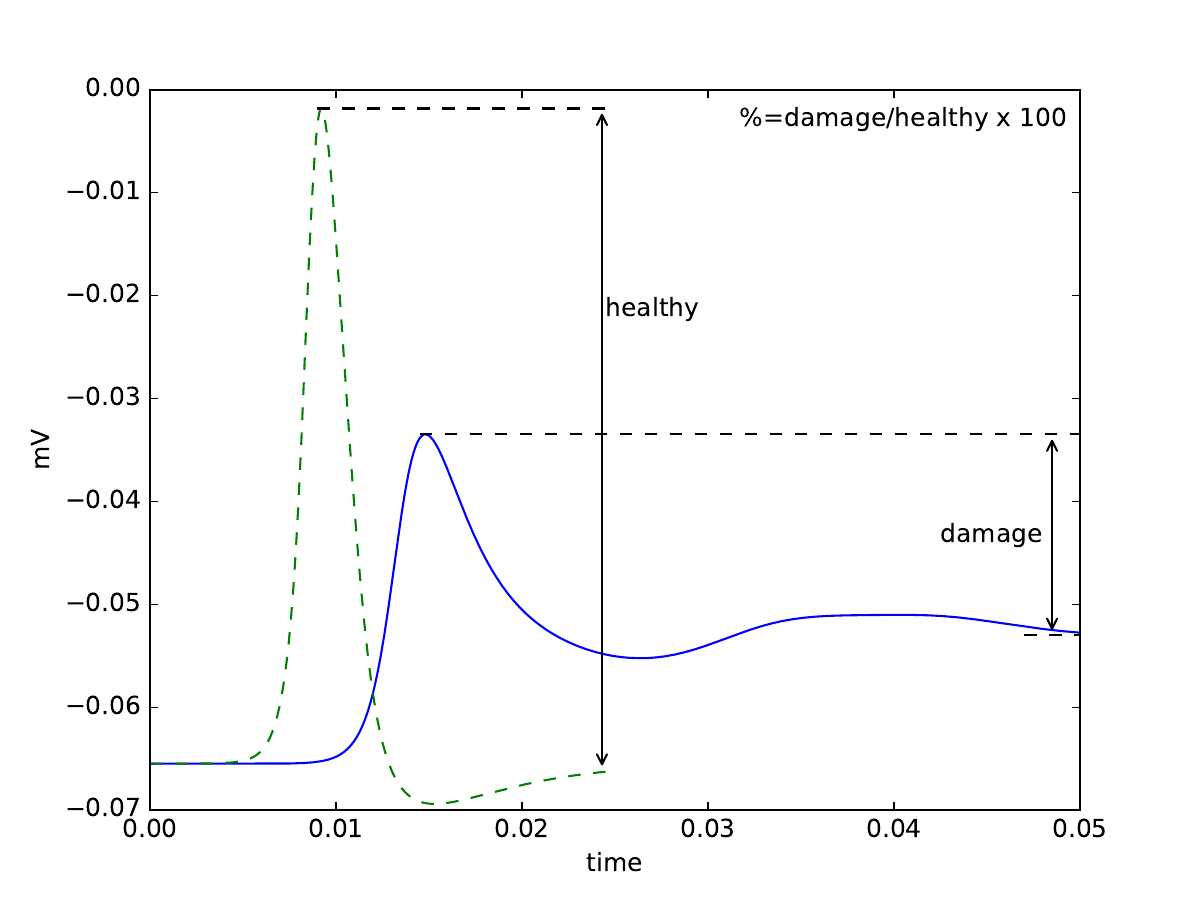}
  \caption{Computation of the fitness value for an arbitrary case.}
  \label{fig:fitness}
\end{figure}

A graphical representation on
how this ratio (\%CAP) is computed is given in Figure \ref{fig:fitness},
where a sample damage case simulation is compared to its reference healthy
counterpart through the following equation:

\begin{equation}
  \%CAP_{ij}=\bigg(\frac{damage_{i,j}}{healthy_{i,j}}*100\bigg)
  \label{eq:capred}
\end{equation}

The model is then contrasted with the measures obtained by the experimental
measurements of the damaged to undamaged CAP ratio propagation
under the same loading conditions \cite{Shi2006}
(see Figure \ref{fig:expData}).

It is worth emphasizing that this original approach proposed in Ref. \cite{Jerusalem2014,GarciaGrajales2015} suffers from two important limitations: i) it only considers one axon, and ii) the signal is triggered by one unique impulse which plays a double role: pushing the potentially unstable resting potential resting potential (due to the damage alteration) towards its new naturally stable  (if there is one) resting potential and providing the signal to be measured for the test (see Figure \ref{fig:fitness} where the tail of the damaged signal is not at the same level as before it is triggered). Ideally, one would want to send a train of signal to change the resting potential, and then only measure the signal. While this will be done in the more realistic bundle scenario in the next section, those two approximations are kept to speed up the simulations and identify the optimization parameters of the optimization framework.

\subsubsection{Optimization problem}
\label{sec:fitnessfunc}
The problem of finding the best parameters values for the electrophysiological-mechanical model
described in Section \ref{sec:em_model} can be formulated as a global continuous optimization
problem. This kind of problems involve the minimization or maximization of a specific
\emph{objective/fitness} function. In the remainder of this paper, we consider the minimization
case, without loss of generality. Mathematically, minimizing an objective function means
finding $x^\text{*}$ defined by Equation \eqref{eq:minimization}:

\begin{equation}
  x^* \in [a,b] \mbox{ where } f(x^*) \le f(x), \forall x \in [a, b]\\
\label{eq:minimization}
\end{equation}

\noindent where $a, b \in \mathbb{R}^N$ and $f:\mathbb{R}^N\rightarrow \mathbb{R}$ 
is called the \emph{objective/fitness} function.

In our problem, fitness was evaluated by the ability of a model parameterization to match the experimental
electrophysiological deficits -- the percentage of CAP reduction due to a 
loading case (six loading cases and seven time points: 42 comparisons in
total) relative to a signal of a non-loaded bundle.

The fitness function, $f$, is defined as the accumulative score for
each time point compared to the experimental reference (see Figure \ref{fig:expData}):

\begin{equation}
  \centering
  f = \sum_{i,j}\left|\%CAP_{ij}^{s}-\%CAP_{ij}^{e}\right|
  \label{eq:fitness}
\end{equation}

{\setlength{\parindent}{0cm}where $i$ represents the different loading cases (listed in Table \ref{table_cases}), $j$ represents the seven time points (from \SI{5}{\minute} time interval up to \SI{30}{\minute}), and where $\%CAP^{s/e}_{ij}$ correspond to the $\%CAP$ of loaded Case $i$ at time $j$ relative to the non-loaded case for the simulation/experiment, see Table \ref{table_cases}. All these simulations are run with the parameters values identified by the optimization algorithm.}

\begin{table}[thpb]
\begin{center}
\caption[Tensile tests loading conditions.]{Tensile tests loading conditions: the six different combinations of strain rate $\dot{\varepsilon}$ and maximum strain $\varepsilon_{\hbox{max}}$ \cite{Jerusalem2014}.}
\label{table_cases}
\begin{tabular}{c c c c}
\hline
Case & Description & $\dot{\varepsilon} (s^{-1})$ & $\varepsilon_{max}$ ($\%$)\\
\hline
1 & Slow-mild & 0.006-0.008 & 25\\
2 & Slow-moderate & 0.006-0.008 & 50\\
3 & Slow-severe & 0.006-0.008 & 100\\
4 & Fast-mild & 355-519 & 25\\
5 & Fast-moderate & 355-519 & 50\\
6 & Fast-severe & 355-519 & 100\\
\hline
\end{tabular}
\end{center}
\end{table}

\subsection{Simulation setup and parameter tuning}
\label{sec:scenario}
Both the DE algorithm and the NEURITE simulator
were coded in C++ and tested with the machine
configuration shown in Table \ref{table:pcconfiguration}.

\begin{table}[!ht]
\begin{center}
\caption{Computer configuration.}
\label{table:pcconfiguration}
\begin{tabular}{lc}
\hline
\multicolumn{2}{c}{Computer configuration}                 \\
\hline
\textbf{PC}               & Intel Xeon 8 cores 1.86Ghz CPU \\
                          & 22GB RAM                       \\
\textbf{Operating System} & Ubuntu Linux 16.04             \\
\textbf{Prog. Language}   & C++                            \\
\textbf{Compiler}         & GNU C++ 5.4.0                  \\
\hline
\end{tabular}
\end{center}
\end{table}

Evaluating one candidate solution (six damage cases and seven
time points) requires approximately 8 min to run for the
computer with the aforementioned configuration in the sequential case. Due to this
computing-time constraint, the parameter tuning of the algorithm
had to be limited in terms of number of the considered configurations,
see Table \ref{table:params}, where the values finally
selected after $25$ independent runs are provided in bold. In particular, two different crossover
operators were tested (binomial and exponential), two population
sizes (preferring the smallest one of the two) and two values
for the weighting factor F and the crossover constant CR \cite{Storn1997}.
An exhaustive grid search was conducted on the eight possible
combinations of these three parameters values.
The final DE strategy used in our simulations is thus the \textit{de/rand/1/exp}.

\begin{table}[!ht]
\begin{center}
\caption{Parameters values for the DE algorithm.}
\label{table:params}
\begin{tabular}{lc}
\hline
Parameter           & Value                          \\
\hline
crossover operator  & binomial, \textbf{exponential} \\ 
population size     & \textbf{15}, 25                \\ 
F                   & 0.1, \textbf{0.5}, 0.9         \\ 
CR                  & 0.1, \textbf{0.5}, 0.9         \\
Fitness evaluations & 450                            \\
\hline
\end{tabular}   
\end{center}
\end{table}

The bottleneck of the optimization
algorithm for this model calibration problem is the fitness function, which
requires the execution of a simulation with the parameters values encoded
in the candidate solution. For this reason, we decided to use the
OpenMP\footnote{https://www.openmp.org/} API to parallelize the computation
of the fitness function. As shown in Equation \eqref{eq:fitness}, computing
the fitness of a solution involves the summatory of the differences between
simulated and experimental CAPs for each of the damage cases and time points.
In our simulations, this means that 42 threads (six damage cases and
seven time points) could eventually run in parallel.

Moreover, DE, as other metaheuristics such as Genetic
Algorithms, is a population-based algorithm, which means that, at each iteration,
it maintains a population of candidate solutions. As solutions can be evaluated
independently, the parallel implementation of DE that we are using can run
$42 \times NP$ processes concurrently, where $NP$ represents the population size
used by the DE algorithm. Of course, the actual number of concurrent threads
will be limited by the available number of cores and we can expect a linear
speedup compared to the sequential execution time reported in the previous section.
In our case, we are limited by the configuration of the hardware platform used
for the experiments to a maximum of $16$ concurrent threads (see Table
\ref{table:pcconfiguration} for details).

\subsection{Results}
\label{sec:res}
The model calibration was conducted in two steps, so as to study
the dependencies between the values of some of the variables. In the first
step, we considered the search space intervals for the free parameters of
the model reported in Table \ref{table:search_params}.

\begin{table}[!ht]
\begin{center}
\caption{Search space for the model's parameters.}
\label{table:search_params}
\begin{tabular}{lrl}
\hline
Parameter             & \multicolumn{2}{c}{Search space} \\
\hline
$E$                   & [$10^3$, $10^6$] & Pa   \\ 
$k$                   & [$10^3$, $10^7$] & Pa   \\ 
$\eta_{eq}$           & [$10^6$, $10^8$] & Pa s \\ 
$\tilde{\varepsilon}$ & [0, 0.4]         &      \\
$\kappa$              & [0, 1]           &      \\
$\gamma$              & [1, 4]           &      \\
\hline
\end{tabular}   
\end{center}
\end{table}

\begin{figure}[!ht]
    \centering

    \begin{subfigure}[b]{0.45\textwidth}
        \includegraphics[width=\textwidth]{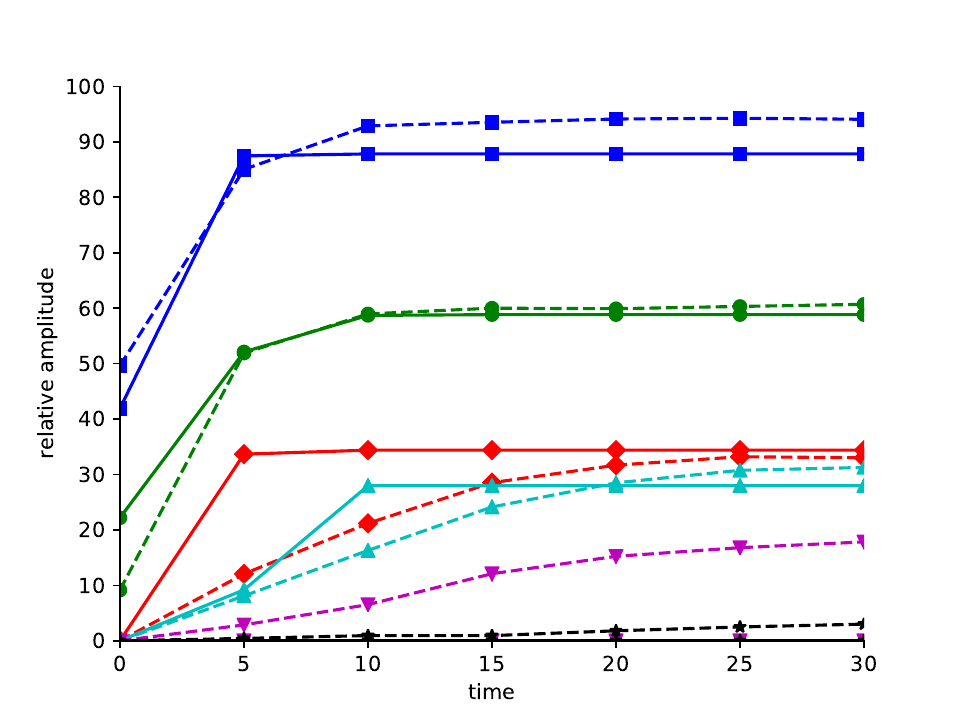}
        \caption{Automatic calibration (original search space for $\eta_{eq}$) vs. simulation results.}
        \label{fig:ref_vs_eta_eq_28}
    \end{subfigure}
    ~
    \begin{subfigure}[b]{0.45\textwidth}
        \includegraphics[width=\textwidth]{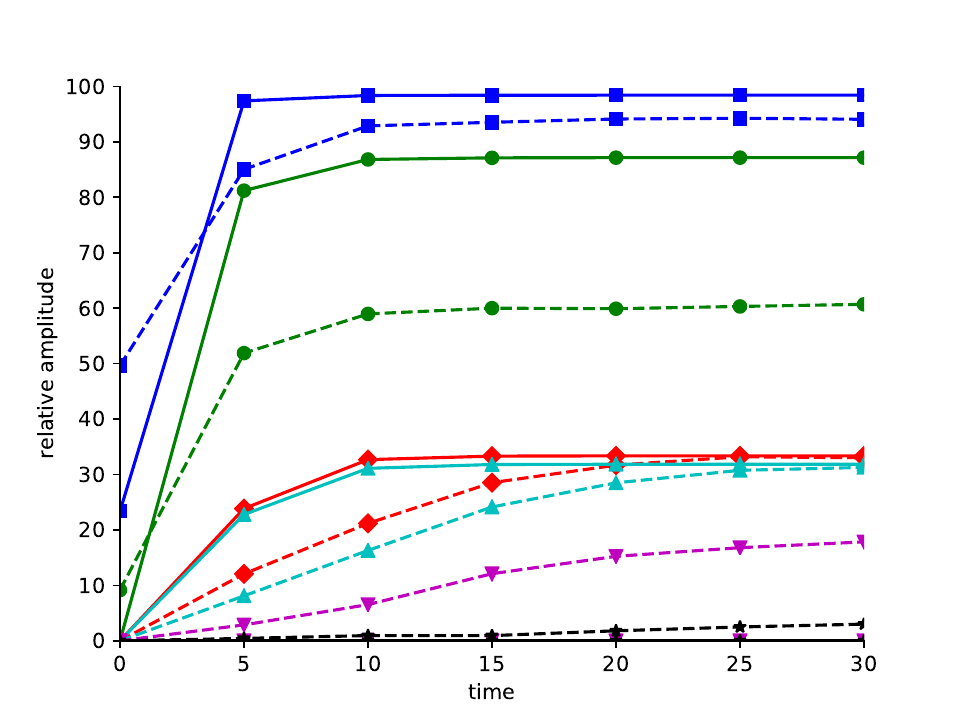}
        \caption{Manual calibration vs. simulation results.}
        \label{fig:ref_vs_manual}
    \end{subfigure}

    \begin{subfigure}[b]{0.70\textwidth}
        \includegraphics[width=\textwidth]{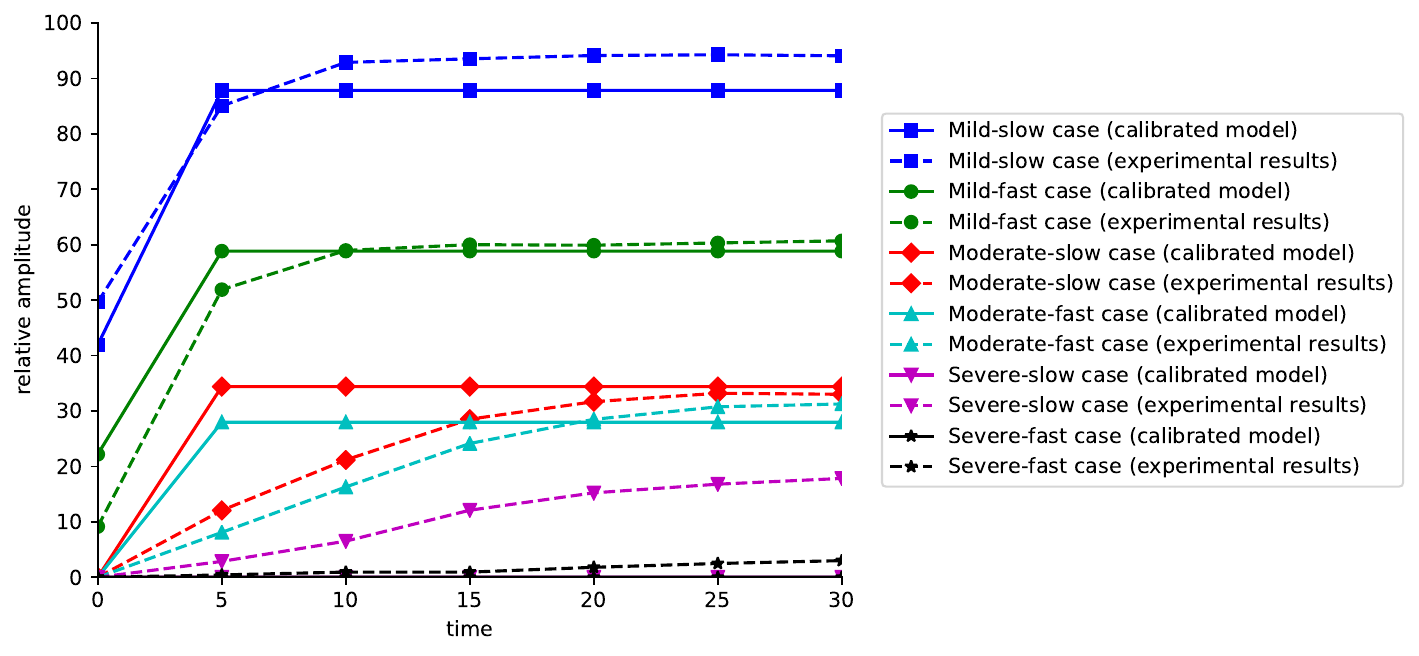}
        \caption{Automatic calibration (reduced search space for $\eta_{eq}$) vs. simulation results.}
        \label{fig:ref_vs_eta_eq_00}
    \end{subfigure}
    \caption{Comparison of manual and automatic calibrations of the model. In each plot, the blue lines
    with squares represent the mild-slow case, green lines with circles the mild-fast case, red lines with diamonds the moderate-slow case, cyan lines with triangles the moderate-fast case, purple lines with
    upside-down triangles the severe-slow case and, finally, black lines with stars the severe-fast case.}
    \label{fig:adj_complete}
\end{figure}

Figure \ref{fig:ref_vs_eta_eq_28} compares the simulation results of a
representative calibration obtained with the DE
algorithm (DE model 1) against the experimental results reported in Ref. \cite{Shi2006}.
As can be seen, the model is able to capture quite accurately the behavior
the mild cases (both slow and fast) and reasonably well the moderate cases
(again, both slow and fast) except at time 10 min for the moderate-fast case
and times 5 and 10 min for the moderate-slow case. If we compare these
results with those of the manual calibration shown in Figure \ref{fig:ref_vs_manual},
we can see how the automatic calibration is able to obtain better results
than the manual calibration for most of the cases. This is especially true for the
mild damage cases and, in particular, for the mild-fast case.
It is interesting to note how the model is still not able to fit
the severe cases. This potentially
arises for the extreme sensibility of the signal collapse for large damage,
see Ref. \cite{GarciaGrajales2015,Jerusalem2014}, and the simplification of the triggering signal.

As can be seen in Table \ref{table:search_params}, the maximum possible value
for the $\eta_{eq}$ parameter is two orders of magnitude larger than
that of the $E$ parameter. This has an important effect on the simulation in
terms of how the steady state can be reached. This can be clearly seen if we
run again the optimization algorithm constraining the search space for the
$\eta_{eq}$ from [$10^6$, $10^8$] to [$10^6$, $10^7$] (DE model 2). The results of this
new scenario are shown in Table \ref{table:full_results}: the DE algorithm
is able to minimize to very low
values the error made. A closer look reveals that, for many of the
damage cases and times, this error is almost zero, as will be shown
graphically later. Compared with the results of the manual calibration \cite{GarciaGrajales2015}
(also shown in Table \ref{table:full_results}) we can observe that even the
worst run of the DE algorithm obtains significantly better fitness than the
manual calibration. Furthermore, the best result of DE is twice as good as
the manually calibrated one.

\begin{table}[!ht]
\begin{center}
\caption{Results of 25 independent executions of DE model 2 vs. manual calibration}
\label{table:full_results}
\begin{tabular}{lrl}
\hline
          & Avg. fitness \\
\hline
Best      & 23.24        \\ 
Median    & 32.67        \\ 
Worst     & 41.77        \\ 
Mean      & 33.05        \\
Std. Dev. &  5.34        \\ \\
Manual    & 52.56        \\
\hline
\end{tabular}   
\end{center}
\end{table}

Figure \ref{fig:ref_vs_eta_eq_00} depicts another representative execution for
this second batch of runs with the constrained search space for the $\eta_{eq}$
parameter of the model. It can be observed
how the model fails then to capture time 5 min for the moderate-fast case and
worsens its ability to fit the same time instant for the mild-fast case.
Table \ref{table:results_params} contains the best results for the manually
and the automatically calibrated models with DE with and
without the search space constraint for the $\eta_{eq}$ parameter (DE models 1 and 2).

\begin{figure}[ht]
  \centering
  \includegraphics[width=1.0\textwidth]{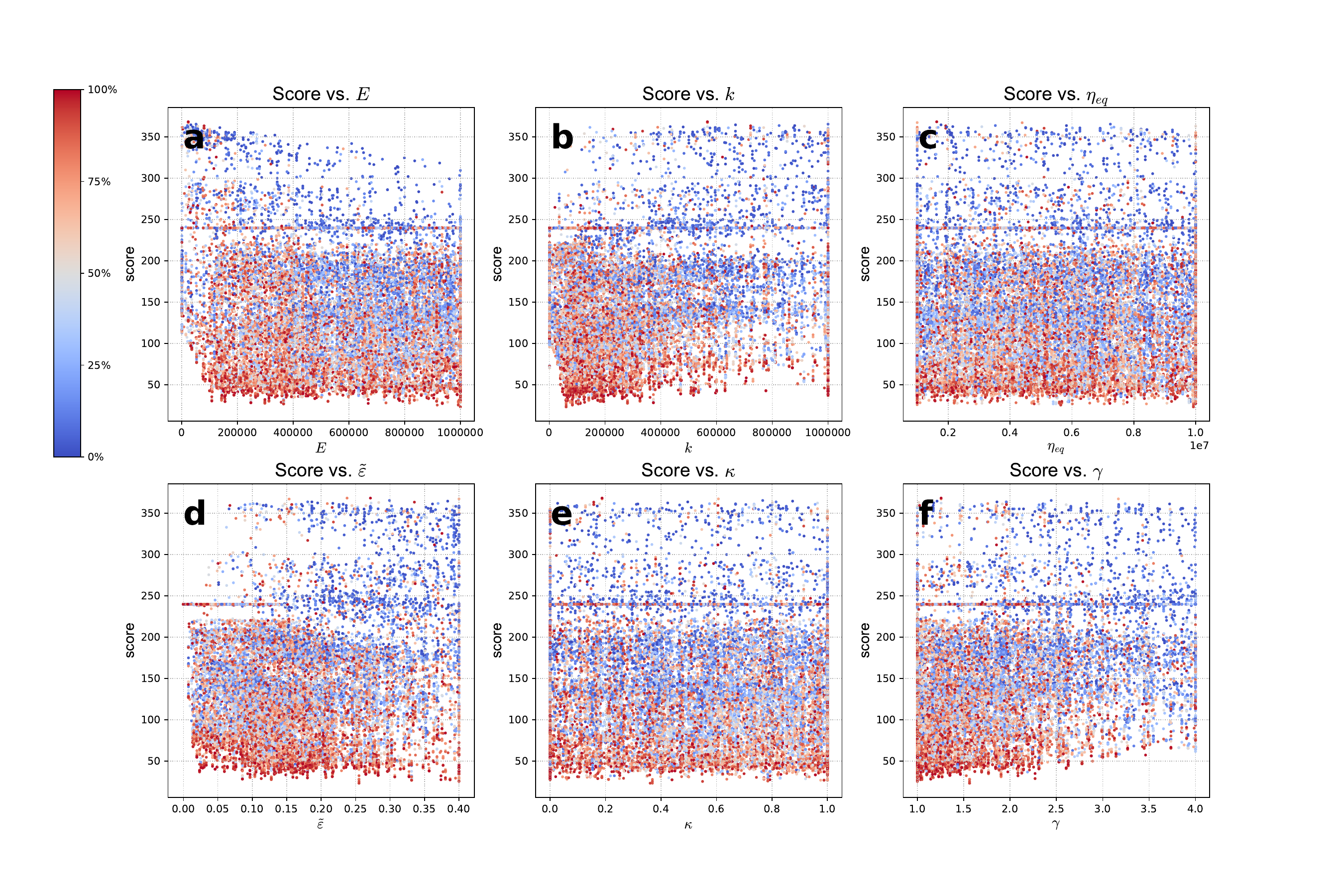}
  \caption{Relationship between parameters and score values obtained for the fitness function. The colour
  encoding represents the instant, in terms of number of iterations, in which
  the value was sampled. Blue dots are values generated in the early steps of
  the evolutionary process whilst red dots are the values that DE generates in
  the last iterations of each run.}
  \label{fig:params_vs_fitness}
\end{figure}

Figure \ref{fig:params_vs_fitness} depicts the relationship between fitness
values and each of the parameters of the coupled mechanical-electrophysiological model.
From this analysis, several observations can be made:

\begin{enumerate}
  \item There does not seem to be any dominant attribute in the fitness
  function. Only $\Gamma$ and $k$ seem to be biased, in both cases, to small
  values of these parameters. On the other hand, what we have instead is a
  high degree of interaction among the parameters (high epistasis). Such
  epistasis is in agreement with the physical interpretation of the free
  parameters optimized here, where each of them is aimed at representing
  the mechanical properties of the axon, see Ref. \cite{Jerusalem2014}.
  \item The optimization algorithm is able to find very low fitness values,
  as mentioned before. The search also progresses during the whole execution,
  which could mean that and additional budget of fitness evaluations would
  allow the algorithm to find even better solutions.
  \item There is a noticeable horizontal line at a fitness value of around 240.
  We hypothesize that this could be a basin of attraction (with a large neutral
  neighbourhood). However, more
  experiments are needed to assess any of those hypotheses.
\end{enumerate}

\begin{table}[!ht]
\begin{center}
\caption{Best configurations for the manually calibrated model (manual model),
automatically calibrated model with DE (DE model 1) and
automatically calibrated model with constrained search space for the $\eta_{eq}$
parameter (DE model 2).}
\label{table:results_params}
\begin{tabular}{llll}
\hline
                      & Manual model                  & DE model 1                    & DE model 2                    \\
\hline
E                     & \SI{1.66e+05}{\pascal}        & \SI{5.62e+05}{\pascal}        & \SI{5.71e+05}{\pascal}        \\ 
k                     & \SI{1.85e+05}{\pascal}        & \SI{2.80e+05}{\pascal}        & \SI{2.17e+05}{\pascal}        \\ 
$\eta_{eq}$           & \SI{1.85e+07}{\pascal\second} & \SI{4.42e+07}{\pascal\second} & \SI{1.00e+06}{\pascal\second} \\ 
$\tilde{\varepsilon}$ & \SI{1.00e-01}{}               & \SI{1.42e-01}{}               & \SI{1.57e-01}{}               \\
$\kappa$              & \SI{5.00e-01}{}               & \SI{6.90e-01}{}               & \SI{2.42e-01}{}               \\
$\gamma$              & \SI{2.00e+00}{}               & \SI{1.00e+00}{}               & \SI{1.08e+00}{}               \\
\hline
\end{tabular}   
\end{center}
\end{table}

\begin{figure}[!ht]
    \centering
    \begin{subfigure}[b]{0.45\textwidth}
        \includegraphics[width=\textwidth]{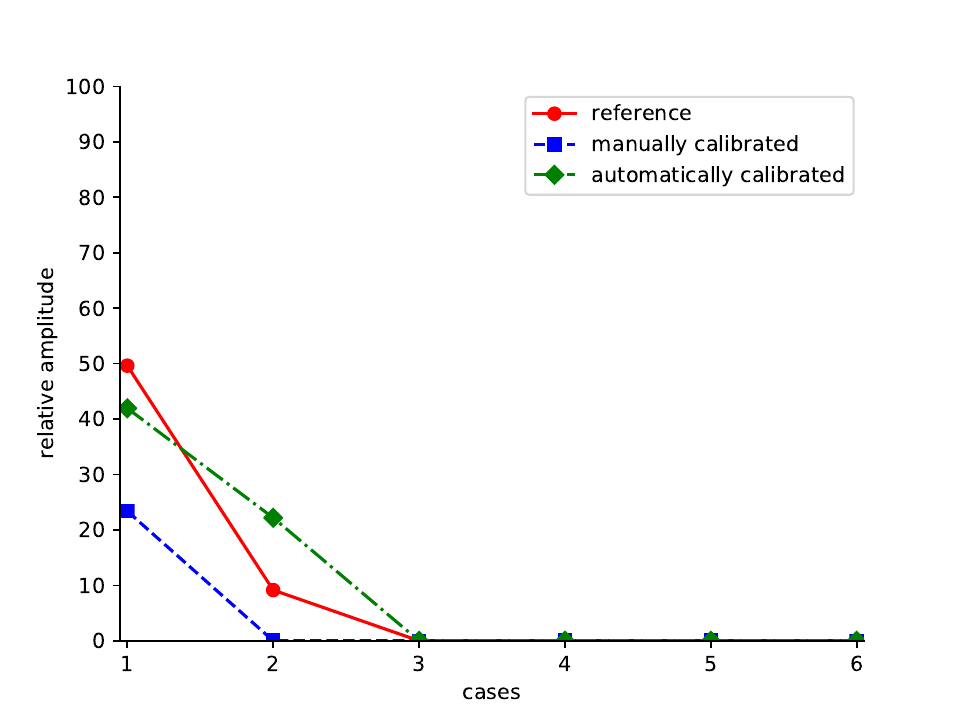}
        \caption{t=0 minutes}
        \label{fig:t0}
    \end{subfigure}
    ~
    \begin{subfigure}[b]{0.45\textwidth}
        \includegraphics[width=\textwidth]{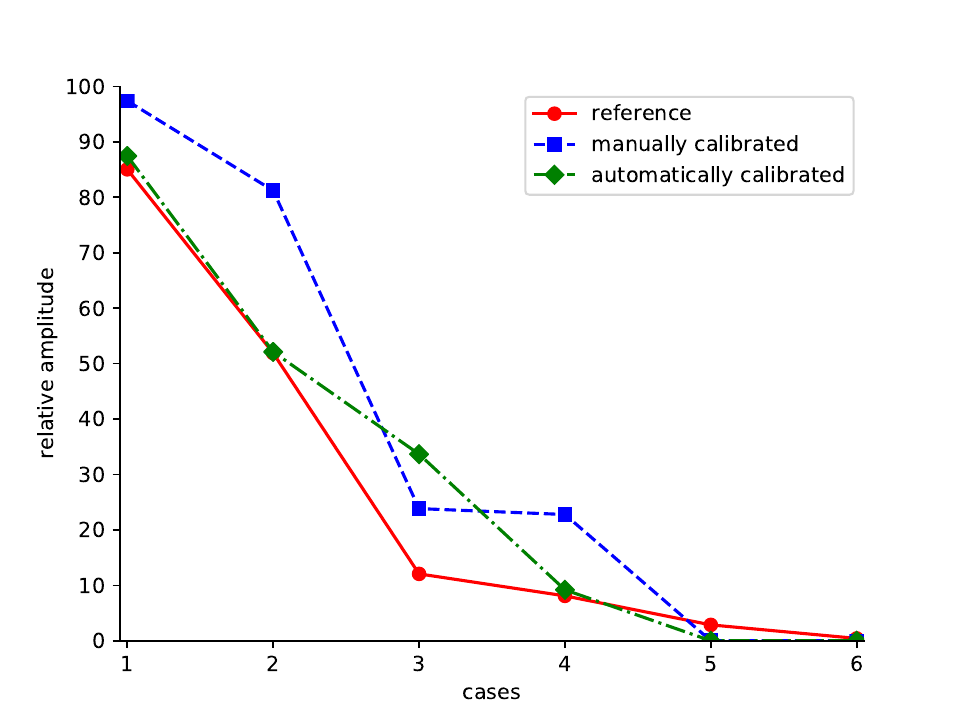}
        \caption{t=5 minutes}
        \label{fig:t5}
    \end{subfigure}

    \begin{subfigure}[b]{0.45\textwidth}
        \includegraphics[width=\textwidth]{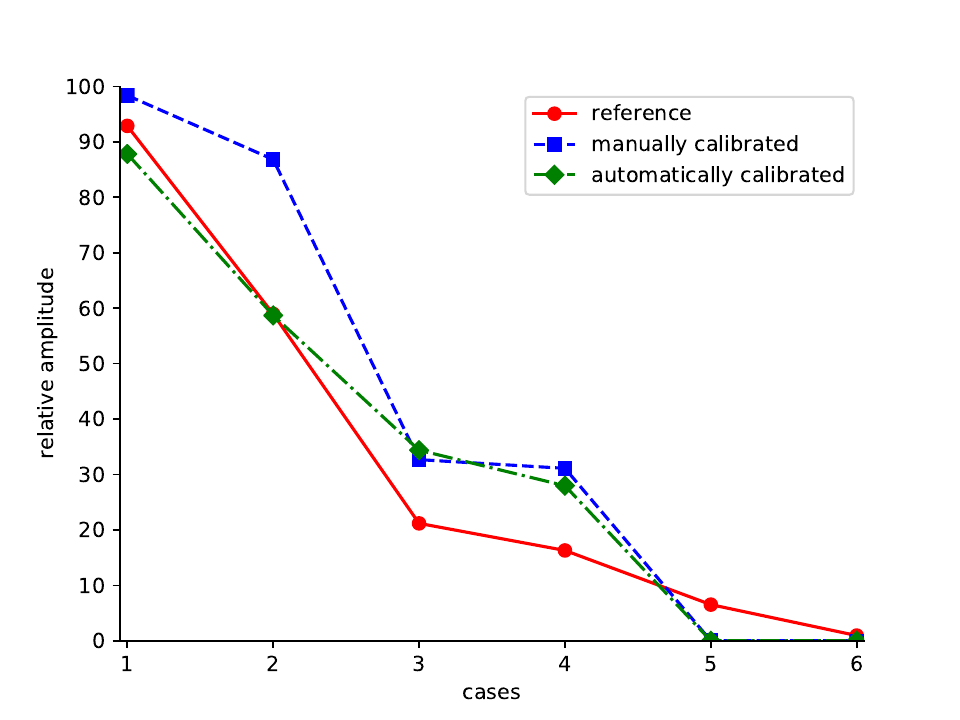}
        \caption{t=10 minutes}
        \label{fig:t10}
    \end{subfigure}
    ~
    \begin{subfigure}[b]{0.45\textwidth}
        \includegraphics[width=\textwidth]{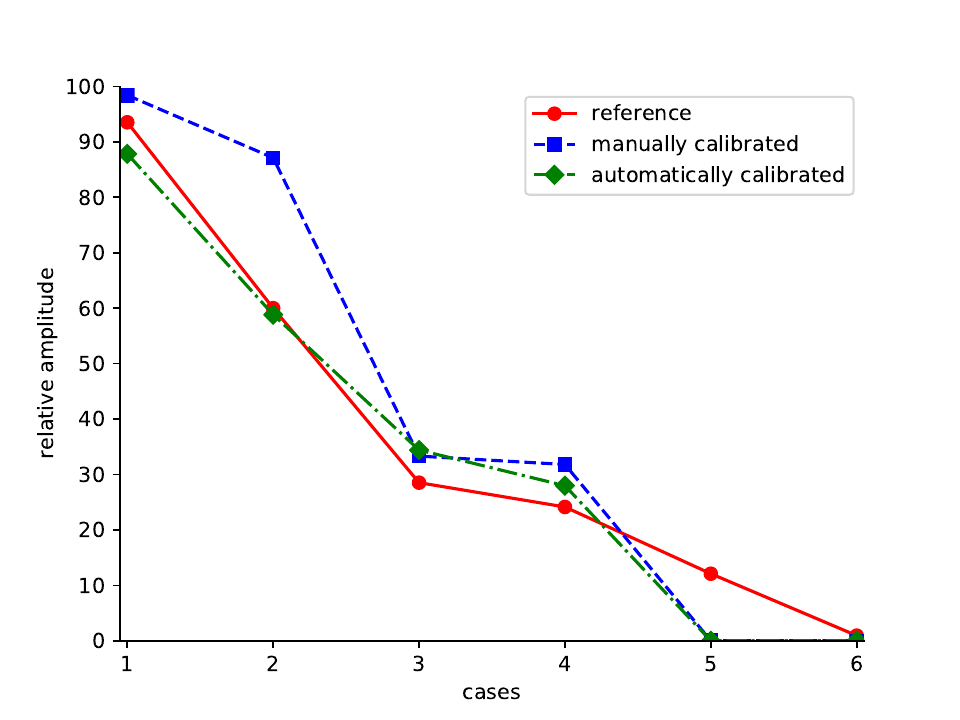}
        \caption{t=15 minutes}
        \label{fig:t15}
    \end{subfigure}
    \caption{Adjustment of the model to the reference calibration
    for each case and instants of times 0-15 min. Red solid line with circles
    represents the reference calibration. Blue dashed line with
    squares represents the manually calibrated model. Green dotted
    line with diamonds represents the automatically calibrated
    model with the DE algorithm. Each plot
    compares the manual and the automatic adjustments with
    regards to the reference calibration for the six damage cases
    and seven instants of time. Each point on the x-axis represents:
    (1) mild-slow case; (2) mild-fast case; (3) moderate-slow case;
    (4) moderate-fast case; (5) severe-slow case; and (6) severe-fast
    case.}
    \label{fig:adj_per_times_1}
\end{figure}

\begin{figure}[!ht]
    \centering
    \begin{subfigure}[b]{0.45\textwidth}
        \includegraphics[width=\textwidth]{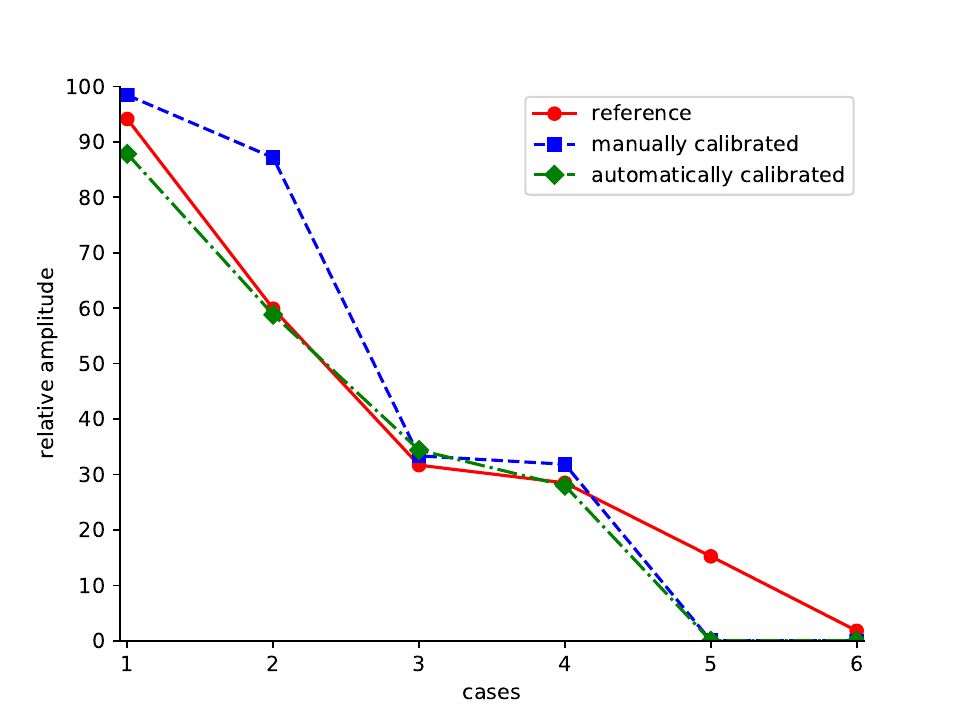}
        \caption{t=20 minutes}
        \label{fig:t20}
    \end{subfigure}
    ~
    \begin{subfigure}[b]{0.45\textwidth}
        \includegraphics[width=\textwidth]{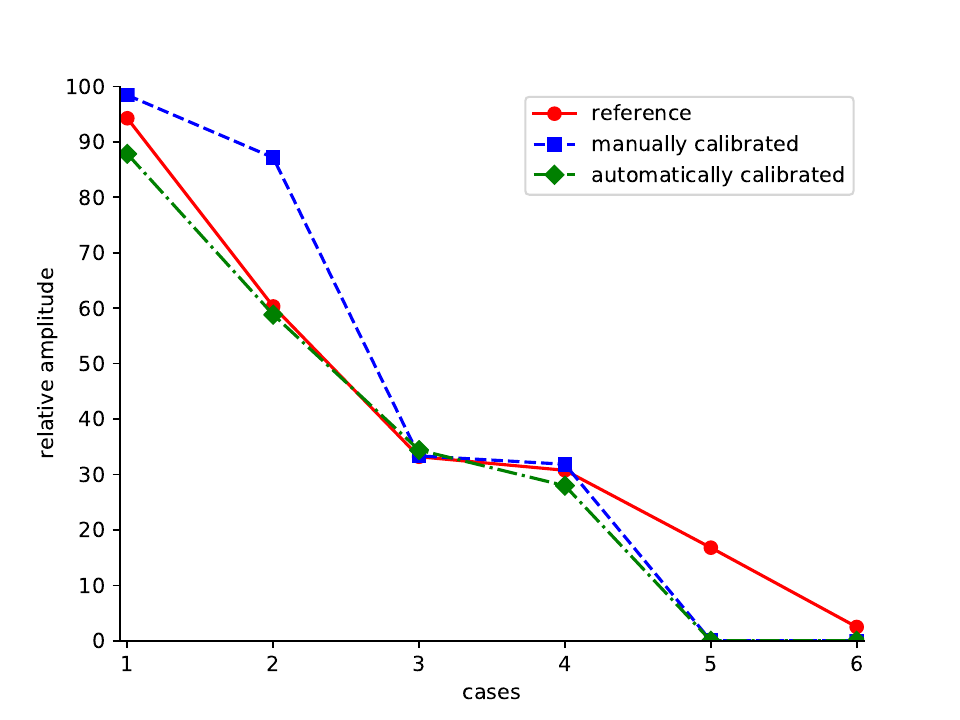}
        \caption{t=25 minutes}
        \label{fig:t25}
    \end{subfigure}

    \begin{subfigure}[b]{0.45\textwidth}
        \includegraphics[width=\textwidth]{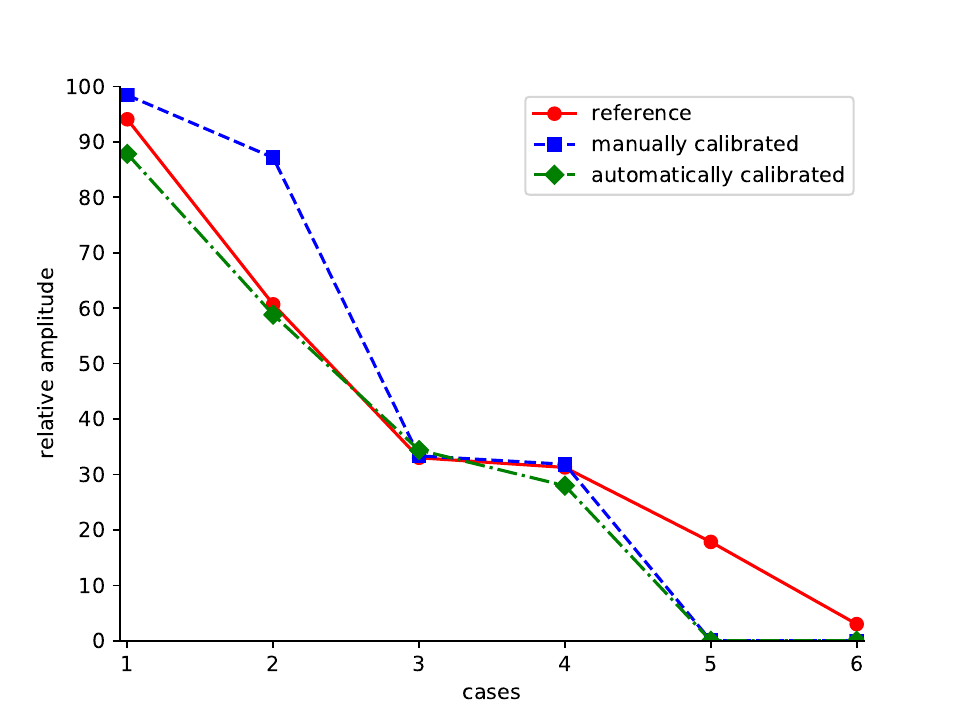}
        \caption{t=30 minutes}
        \label{fig:t30}
    \end{subfigure}
    \caption{Adjustment of the model to the reference calibration
    for each case and instants of times 20-30 min. Red solid line with circles
    represents the reference calibration. Blue dashed line with
    squares represents the manually calibrated model. Green dotted
    line with diamonds represents the automatically calibrated
    model with the DE algorithm. Each plot
    compares the manual and the automatic adjustments with
    regards to the reference calibration for the six damage cases
    and seven instants of time. Each point in the x-axis represents:
    (1) mild-slow case; (2) mild-fast case; (3) moderate-slow case;
    (4) moderate-fast case; (5) severe-slow case; and (6) severe-fast
    case.}
    \label{fig:adj_per_times_2}
\end{figure}

Figures \ref{fig:adj_per_times_1} and \ref{fig:adj_per_times_2} provide
an alternative perspective on the
simulation results by comparing the results of the reference calibration
with those of the manual and automatic calibrations. In these plots, there
are several things that should be highlighted. First, the automatic calibration
is able to reach an almost perfect
fit for both mild cases for all the time instants, except at time 0 min. Second,
the automatic calibration obtains better results for all the cases and times
except for mild-fast damage case at time 0 min and the moderate-slow at
all the time points.
Finally, none of the calibrations is able to fit the severe damage cases.
This discrepancy might find its roots in the following: i) only one axon was used in the simulation setup (as opposed to the setup that makes use of a bundle of axons), and ii) as discussed earlier, the electrical signal is triggered without a preconditioning of the new axon, i.e., the damage might lead to a new resting potential that can only be reached by first disturbing the axon; after which, the signal to be measured can be sent; in the previous setup, only the latter was done.
Those two points are tackled in the following.

\subsection{Comparison with other optimization algorithms}
\label{sec:cmp_opt}

To assess the performance of DE as the optimizer
for our problem, we have conducted a comparison with two other methods:
(1) a gradient-based optimizer, such as the Broyden–Fletcher–Goldfarb–Shanno
(BFGS) algorithm \cite{Fletcher1994}; and (2) a more advanced DE-based algorithm,
the Success History based DE with Linear Population Size Reduction (LSHADE)
\cite{Tanabe2014}. The objective of this comparison is two-fold. On the one hand,
we want to ensure that a population-based metaheuristic needs to be used and that
a simpler gradient-based method is not powerful enough for this complex problem.
On the other hand, we want to check if a more sophisticated algorithm such as
LSHADE can yield even better results than DE. To allow for a fair comparison, we have
run both algorithms for the same number of maximum fitness evaluations as DE.

Table \ref{table:cmp_results} shows the results obtained by these two other
optimization methods. The results of the
DE algorithm already reported in Table \ref{table:full_results}, are also included.

\begin{table}[!ht]
\begin{center}
\caption{Results of 25 independent executions of DE, LSHADE and BFGS.}
\label{table:cmp_results}
\begin{tabular}{lrrr}
\hline
          & DE    & LSHADE & BFGS   \\
\hline
Best      & 23.24 & 42.57  &  62.46 \\
Median    & 32.67 & 55.13  & 157.38 \\
Worst     & 41.77 & 79.72  & 350.60 \\
Mean      & 33.05 & 57.07  & 179.40 \\
Std. Dev. &  5.34 & 10.56  &  90.80 \\
\hline
\end{tabular}   
\end{center}
\end{table}

The results of the BFGS algorithm are much worse than those of DE,
as one could expect for a complex problem such as this one. The fitness landscape
of this problem is probably too convoluted for a gradient-based method to work.
Regarding the results of LSHADE, these are not as bad as those of BFGS but are still
worse than those of DE. In this case, the convergence
of LSHADE is a priori slower than that of the classic DE, as it needs to self-adjust several
parameters based on the information stored in the archive, and the limited number
of fitness evaluations allowed in our experiments is too small for the algorithm
to obtain enough information for this self-adjustment to take place. This is no surprise, as
we have previously observed this behaviour in other self-adaptive algorithms \cite{LaTorre2010a,LaTorre2010b}.

Given that the axonal bundle study will share most of the characteristics of the
single axon problem, we will use DE as the optimizer of choice for the remainder of
this paper.


\section{Axonal bundle study}
\label{sec:axon2nerve}
In Section \ref{sec:singleaxon}, the model by Jerusalem et al. \cite{Jerusalem2014} of the altered axonal electrophysiology caused by mechanical stretch was wrapped within a parallel DE optimization framework to obtain a calibration of its parameters; outperforming the manual calibration originally carried out in Ref. \cite{Jerusalem2014}. It is however unclear whether the overall CAP of a segment of spinal cord tissue post-stretch (as measured experimentally by Shi et al. \cite{Shi2006}) can be modelled by the AP of one unique axon (such as the model by Jerusalem et al. \cite{Jerusalem2014}). Additionally, the axonal resting potential requires first a rebalancing before the transmitted signal can be tracked. While this can simply be done by sending first a signal to obtain the new resting potential before sending the signal to be tracked, this was avoided in both the manual and automatic calibration presented above. This section therefore sets out to explore the modelling of altered macroscopic electrophysiology of a bundle of axons in a more realistic way.

Section \ref{art_bundle} proposes an extension framework for extrapolating axon bundle electrophysiology from individual axon electrophysiology, based on the mechanical-electrophysiological single axon model  used earlier. The extended framework is then linked with the previously tuned parallel DE algorithm to perform systematic parameter search to obtain the best calibration against the experimental reference.

\subsection{Axonal bundle model}
\label{art_bundle}
The previously calibrated model, as published by Jerusalem et al. \cite{Jerusalem2014}, uses a reference axon diameter of \SI{3}{\micro\metre} to capture tissue level electrophysiology. However, wide ranging geometrical variations are observed within the axon population of a fibre \cite{Perge2012, Lascelles1966, Vizoso1950, Fride1980, Fride1985a, Fride1985b, Fride1986}. The characteristics of one AP are reflective of the conduction properties of the axon of interest, in turn dependent on its geometrical features \cite{Rushton1951}, such as diameter, nodes of Ranvier sizes or internodal lengths. Previous morphometric studies of axons observed variances in all the aforementioned geometrical parameters \cite{Lascelles1966, Vizoso1950, Fride1980, Fride1985a, Fride1985b, Fride1986}. Furthermore, a normal distribution was observed in axon calibres for different species, such as rats and guinea pigs \cite{Perge2012, Shi2000}. The g-ratio (the ratio of inner axonal diameter to the total outer diameter \cite{Rushton1951,Chomiak2009}) is a good indicator of axonal myelination and when assuming a constant inner axonal diameter and myelin layer thickness, the number of myelin layers can be estimated. This ratio can vary slightly from species to species, and also from healthy to injured axons \cite{Chomiak2009, Ouyang2010, Sun2012}. In terms of mechanically induced electrophysiology deficits, it is desirable to study the relationship between a population's geometrical characteristics and one hypothetical representative characteristic. The need for a collective behaviour within an axonal bundle may hypothetically be rationalised through the need for signal redundancy. The rest of this section therefore details a framework for extending \textit{NEURITE} up to an axonal fibre model, by taking into account realistic geometrical characteristics.

In order to create the fibre model, morphometric relationships for guinea pig axonal calibre, internodal length and number of myelin layers were obtained from Hilderbrand et al.\cite{Hilderbrand1985}.

\begin{figure}[htbp]
	\centering
  \begin{subfigure}[b]{0.75\textwidth}
    \includegraphics[width=\textwidth]{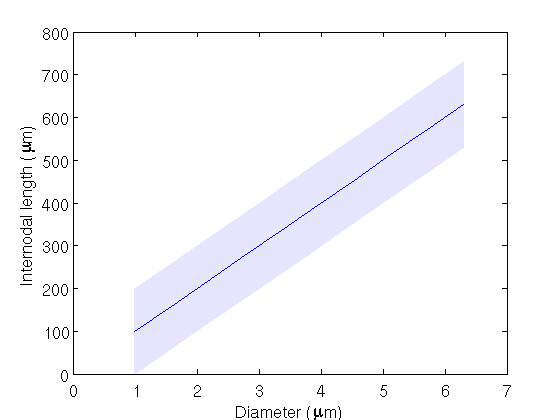}
    \caption{Internodal length vs. diameter relationship}
    \label{ILdia}
  \end{subfigure}
  ~
  \begin{subfigure}[b]{0.75\textwidth}
    \includegraphics[width=\textwidth]{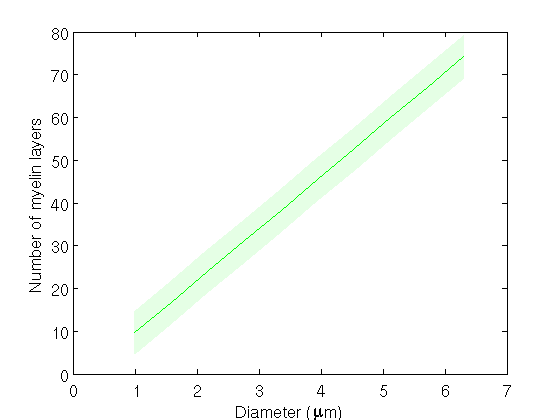}
    \caption{Number of myelin layers vs. diameter relationship}
    \label{ILdia2}
  \end{subfigure}
  \caption[Geometrical relationships extracted from morphmetric studies of guinea pig axons.]{Geometrical relationships extracted from morphmetric studies of guinea pig axons \cite{Hilderbrand1985}.}
\end{figure}

Linear fits were used to approximate the relationship between internodal lengths and the number of myelin layers, and between internodal lengths and axon diameter, see Figure \ref{ILdia} and \ref{ILdia2}.

The internodal length was estimated to be 100 times the axon calibre \cite{Hilderbrand1985}. The smallest diameter in the fibre model was estimated to be \SI{0.7}{\micro\metre} \cite{Perge2012}, and therefore the smallest internodal length was determined to be \SI{70}{\micro\metre}. The number of myelin layers were estimated to be 12 times the diameter value of the axon calibre in micrometres (the slope in Figure \ref{ILdia2})\cite{Hilderbrand1985}.

\begin{figure}[thpb]
  \centering
  \includegraphics[width=\columnwidth]{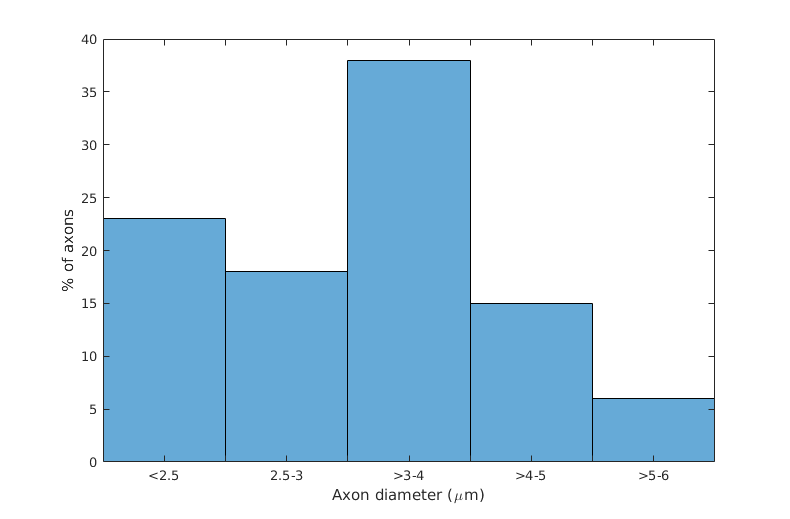}
  \caption[Distribution of diameters in the nerve fibre obtained from guinea pig spinal cord axons.]{Distribution of diameters in the nerve fibre obtained from guinea pig spinal cord axons \cite{Shi2006}.}
  \label{diaDist}
\end{figure}

In the experiments conducted in a double sucrose gap \cite{Shi2006}, the spinal cord segment were over-stimulated with a large voltage, therefore one assumption was that the axons recruited in the CAP reflects the axon population within the cord segment. There are numerous axons within the experimental sample, and it would be impractical and computationally expensive to simulate all of them. Figure \ref{diaDist} presents the reported diameter distribution for the guinea pig spinal cord \cite{Shi2006}. 

In order to simplify the model to a representative population, the data from Figure \ref{diaDist} were taken to extract the relative ratios of the different diameter bins in the histogram. One set of diameters was used and a random diameter within each bin was chosen by a random number generator. Each internode was assigned a length and a number of myelin layers within the range corresponding to its diameter generated by a random number generator. Two axons were used to approximate the smallest histogram bin (axon diameter of 5-\SI{6}{\micro\metre}), and the number of axons per bin for the other bins was chosen proportionally. The minimum number of axons to approximate this population was hence determined to be 27: six axons under \SI{2.5}{\micro\metre}, five between 2.5-\SI{3}{\micro\metre}, ten between 3-\SI{4}{\micro\metre}, four between 4-\SI{5}{\micro\metre} and two between 5-\SI{6}{\micro\metre}. Random diameters within each bin were generated to form the bundle, each with 200 nodes and internodes following the geometrical relationship mentioned above.

\begin{figure}[thpb]
\centering
\includegraphics[width=\columnwidth]{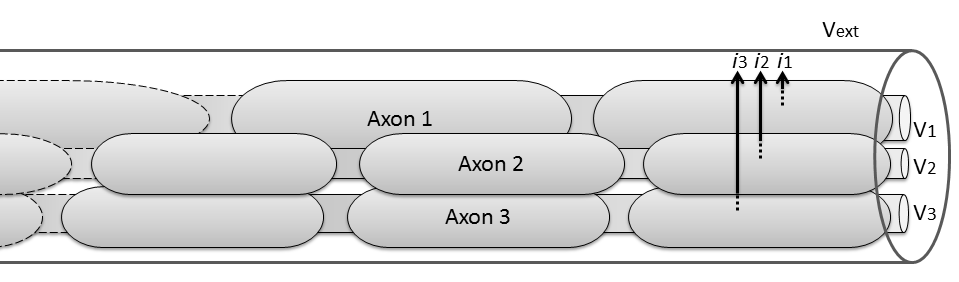}
\caption[Nerve model.]{Nerve model.}
\label{axEq}
\end{figure}

\paragraph{Scaling rule for CAP}\label{sec:scalingRule}
The AP of each axon is obtained at the same point, at \SI{10}{\milli\metre} from the start of the axon. The resultant APs are then averaged using the following scheme to interpolate to an overall signal of a fibre. The contribution of a single axon towards an overall CAP was assumed to be proportional to its circumferential area. This assumption was based on the fact that the ion channels reside on the membrane. If the density of ion channels per unit membrane area is constant, larger axons will produce a larger current, hence resulting in a larger contribution to the overall CAP \cite{Li2007}. This simple scaling rule was adopted for this study and the axon's AP was assumed to be proportional to its diameter, see Figure \ref{axEq}.

For the first single axon (e.g., axon 1 of Figure \ref{axEq}) that is stimulated by an input voltage, the resultant potential measured locally is $V_1$ is such that
\begin{equation}
V_{ext}-V_1 = i_1R_1
\end{equation}and for the $n^{th}$ axon,
\begin{equation}
V_{ext}-V_n = i_nR_n
\label{1}
\end{equation}
where $V_{ext}$, $i_n$ and $R_n$ are, respectively, the extracellular potential, the current leaving axon $n$ and its instantaneous resistance in the vicinity of the measurement point.

\begin{figure}[thpb]
\centering
\includegraphics[width=\columnwidth]{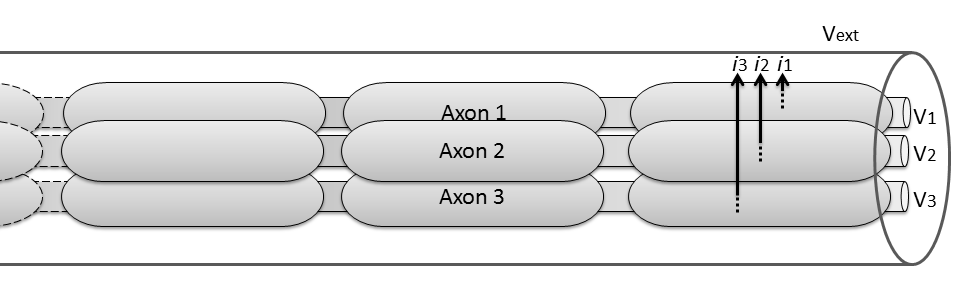}
\caption[A case with three identical axons.]{A case with three identical axons.}
\label{idealCase}
\end{figure}

In a double sucrose gap, and applying the analogy of parallel resistors, the following can be obtained:
\begin{equation}
  V_{ext}-V_{CAP} = i_{Total}R_{Total}
  \label{eq:Vext}
\end{equation}
where
\begin{equation}
R_{Total} = \frac{1}{\frac{1}{R_1}+\frac{1}{R_2}+\frac{1}{R_3}+...+\frac{1}{R_n}}
\label{R}
\end{equation}
\begin{equation}
i_{Total} = i_1 + i_2 + i_3 +...+ i_n
\end{equation}
By using Equation \eqref{1}, Equation \eqref{eq:Vext} becomes:
\begin{equation}
  V_{ext}-V_{CAP} = V_{ext}-\left(\frac{1}{\sum_j^n \frac{1}{R_j}}\sum_i^n \frac{V_i}{R_i}\right)
\label{vCAP}
\end{equation}
As the resistances $R_i$ are related to the patches of membrane in the vicinity of the measurement point, and therefore by definition inversely proportional to the axon diameters $d_i$, this leads to:
\begin{equation}
V_{CAP} = \frac{1}{\sum_j^n d_j}\sum_i^n V_id_i
\label{vCAPd}
\end{equation}
Accordingly, the larger the axon, the higher its contribution towards $V_{CAP}$.
For example, an ideal case of three identical axons all with the same geometry (see Figure \ref{idealCase}) will therefore have the same resistance, and
\begin{equation}
V_{CAP} = \frac{1}{3}(V_1+V_2+V_3)
\end{equation}

This method of scaling may seem straightforward for axons with the same morphology (identical APs), however the assumption of identical morphology is far from reality in biological nerves. APs will differ between axons of different morphologies; furthermore, the same level of strain will affect axons of different calibre differently. The following sections will explore the modelling of electrophysiological alteration at various levels of strain and strain rate based on the above scaling rule.

\subsection{Calibration of axonal fibre model}
\label{sec:exp_nerves}
The fibre model that was evaluated in this simulation is made up of 27 axons and the aforementioned scaling framework is used to produce a CAP. In order to compare the effectiveness the 27-axon bundle, the simulated fibre had a length of \SI{10}{\milli\metre}, the same as for the simulations presented by Jerusalem et al. \cite{Jerusalem2014}. The electrophysiological parameters of each axon was assumed to be the same and the parameters reported in Jerusalem et al. \cite{Jerusalem2014} were used in this study.

To trigger the APs, three voltage stimuli of duration \SI{3}{\milli\second} were used in this study (as opposed to a single stimulus as used by Jerusalem et al. \cite{Jerusalem2014} and used earlier for the optimization of the DE parameters). First, the fibre was triggered twice in quick succession one after the other (at time t=\SI{20}{\milli\second}  and t=\SI{33}{\milli\second}), then at t=\SI{150}{\milli\second} the fibre was triggered again. The reason to adopt this pattern was that it was often observed that a new resting potential can be established for some of the loading cases (especially Cases 3 and 4, see Table \ref{table_cases}). Therefore the first two triggers were used to establish a new stable resting potential before an AP was recorded. Without this triggering system, post stretch CAP for Cases 3 and 4 can have abnormally large amplitudes, since it is possible for the damaged nerve to establish a new resting potential. From a model point of view, allowing the Hodgkin-Huxley model to first establish a new resting potential by triggering it, then triggering it again for an AP is closer to what is measured experimentally. The total simulation time was \SI{300}{\milli\second}, this was chosen to be long enough to observe all APs initiated by the three triggers. Note that $V_{rest} = $ \SI{-65}{\milli\volt} in all simulations in this study.

Fitness is computed analogously as for the one-axon case (see Section \ref{sec:fitnessfunc}). As for the parallelization of the computation of the fitness function, having multiple axons adds an additional level of parallelization: 42 cases/time points combinations, 27 axons and $NP$ independent solutions.

\subsection{Results}
\label{sec:exp_scenario_nerves}
The simulation for the calibration of the axons bundle is carried out on the same computer of the one-axon model, with the configuration discussed in Section \ref{sec:scenario}. With regards to the parameters of the parallel DE algorithm, we have used the values reported in Table \ref{table:params}, which were tuned for the one-axon case.

\begin{figure}[thpb]
  \centering
  \includegraphics[width=140mm]{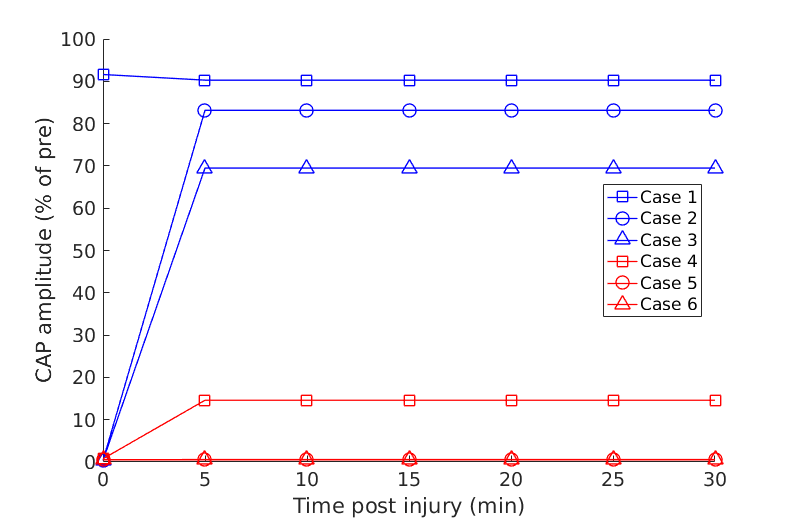}
  \caption[Compound action potential recovery simulations using 27-axon fibre model.]{CAP recovery simulations using 27-axon fibre model.}
  \label{fig:optRepFibre} 
\end{figure}

Figure \ref{fig:optRepFibre} presents results of the 27-axon the model calibrated by the optimization algorithm plotting (CAP reduction vs. time). The calibrated parameter values are provided in Table \ref{table:cal1}. At t=\SI{0}{\minute}, the bundle model experiences a total conduction block except for the mildest strain at the slowest strain rate (Case 1). The CAP for Case 1 at t=\SI{0}{\minute} was predicted to be just over 90$\%$ of  the pre-stretched CAP. While this data point is quite similar to the experimental value presented in Figure \ref{fig:expData}, the CAP at the subsequent relaxation time is higher, which is contrary to the experimental trend. The reason is that the strain was at its highest at t=\SI{0}{\minute} before the relaxation has begun. The axons with diameters between \SI{2.1}{\micro\metre} and \SI{5.8}{\micro\metre} (22 out of the 27 axons) used to build the axon bundle signal become ectopic due to the alteration in the reversal potentials under mild strain and slow strain rates. The ectopic conduction produced in this model was a result of the altered ion channels reversal potential based on membrane strain. If the new reversal potentials cannot establish a resting potential, then the conduction evolves to be ectopic. This effect is then alleviated as the axon enters relaxation, and therefore the CAP for Case 1 from t=\SI{5}{\minute} onwards becomes slightly lower. For the other loading cases however, the maximum loading at t=\SI{0}{\minute} is far too severe, therefore resulting in total conduction block instead of ectopic conduction. As the fibre bundle relaxes, at t=\SI{5}{\minute} onwards, the fibre regains some of its conduction ability for Cases 2, 3 and 4.

The axon bundle model predicts total conduction block for moderate and severe strain at fast strain rate (Cases 5 and 6, respectively), where axons can not recover conduction even after \SI{30}{\minute} of stress free relaxation. This is different to the experimental data where the tissue was still able to recover $\approx20\%$ of CAP for Case 5 and $\approx5\%$ for Case 6.

\begin{table}[thpb]
  \begin{center}
  \caption[Calibrated model parameters values for axon bundle model.]{Calibrated model parameters values for axon bundle model.}
  \begin{tabular}{l l}
    \hline
    Parameter & Value \\
    \hline
    E                     & \SI{4.91e+05}{\pascal}   \\
    k                     & \SI{9.69e+05}{\pascal}   \\
    $\eta_{eq}$           & \SI{5.56e+06}{\pascal s} \\
    $\tilde{\varepsilon}$ & \SI{5.90e-02}{}          \\
    $\kappa$              & \SI{1.56e-02}{}          \\
    $\gamma$              & \SI{2.16e+00}{}          \\
    \hline
  \end{tabular}
  \label{table:cal1}
  \end{center}
\end{table}

Compared to the original experimental data and the one-axon model, it can be seen that the behaviour of the 27-axon fibre model is less gradual. The fibre model exhibits more conduction blocks during the relaxation time, and fails to capture the transient phase (t=5 - \SI{10}{\minute}) in the recovery that is present in the experimental data.

In healthy nerve tissues, axons of different calibres exist to conduct electrophysiological signals. Larger and smaller axons have different conduction speeds and this allows information to be encoded temporally in the signals \cite{Perge2012}. This study has theoretically demonstrated that this heterogeneity may not only serve a purpose in healthy nerves, but also in terms of protective mechanisms of nerves when subjected to mechanical damage. The redundancies and heterogeneity included in the fibre model provided more resilience to the CAP conduction.

The three-trigger approach ensured that the axon was at a new settled resting potential before measuring the CAP loss, while the original model (and one-axon model used earlier for the calibration of the DE parameters) took an unstable resting potential as a reference, leading to biased results. This new resting potential is the result of the altered reversal potentials of the ion channels of the Hodgkin-Huxley model. Through the dependence of reversal potential on the membrane strain (similarly observed by Boucher et al. \cite{Boucher2012}), it was chosen here to exclude the effect of the new resting potential by using three triggers. Since only relative CAP loss values were reported, it was assumed that new resting potentials could often be established experimentally.

The fibre model study shows that it is complicated to represent altered electrophysiology on a higher length scale in terms of linear combination of altered axonal electrophysiology. One possible explanation is that the overall altered CAP is not a linear combination of the altered axonal AP. For example, demyelination is commonly observed as a secondary injury hallmark, and it has sometimes been observed as early as \SI{30}{\minute} \cite{Ouyang2010}, potentially introducing complex signal attenuation and time delays in propagation within a nerve bundle. Another damage mechanism that could occur within the \SI{30}{\minute} recovery window could be the influx of $Ca^{2+}$, which alters the functioning of the $Na^{+}$/$Ca^{2+}$ exchanger \cite{Wang2012}, therefore increasing the complexity of the ion imbalance of $Na^{+}$ and $K^{+}$, contributing to reasons why the altered electrophysiology for one axon, as well as a bundle of axons, may not be straightforwardly modelled by the damage model used in \textit{NEURITE}. Despite the shortfall of the above models in terms of secondary ion dynamics, it is important to acknowledge that the consideration of the axon population redundancies has a beneficial effect on capturing the electrophysiological coupling. The fibre model would have the capability to model a more graded response as some of the smaller axons become more damaged electrophysiology under a certain strain and strain rate. Therefore it may be advisable to consider a fibre model when studying and modelling axonal tracts.

Other than a potentially indescriptive electrophysiological model, it is also possible that the micromechanics model proposed by \textit{NEURITE} may not be capturing the mechanical damage and recovery that is present in an axon within a tissue segment. The mechanical model in \textit{NEURITE} was formulated with the infinitesimal strain theory, while the macroscopic and microscopic levels were linked with a viscous dashpot. It is not straightforward that this is the mechanical relationship between the two length scales, furthermore, it is not yet feasible to validate this kind of multiscale model in biological tissues.


\section{Conclusions}
\label{sec:conclusions}
In this paper, we have used a parallel DE algorithm for
the automatic calibration of the parameters of an electrophysiological simulation
model of the stretch induced axonal deficit.
The model was then extended to consider more complex scenarios
involving multiple axons in the form of fibres and more realistic triggering processes. The same optimization
algorithm was used to calibrate the parameters of the extended model.

The calibration of the single axon model was able to adjust quite effectively for the two
cases of mild mechanical damage, whereas in the case of moderate mechanical
damage the fitness was not as good. The worst cases appear under severe
mechanical damage (both slow and fast). The ability for the model to represent
these extreme scenarios may also be limited by additional effects occurring
experimentally and not captured by the model itself. These effects may
include membrane post-traumatic blebbing or membrane permeability disruption or even
the collapse of the neurite cytoskeleton. All these conditions would be expected to
cause an almost null signal transmission.  While the fibre model was aimed at providing a more realistic prediction, it appears to perform worse than the one-axon model. Again, we believe that this is an unfortunate consequence of our attempt to capture the redundancy and heterogeneity of a real experimental setup. More work is clearly needed.

Finally, it is worth emphasizing that the calibration of a
large number of these elements implies a computationally intensive simulation in
the core of the optimization (calibration) algorithms. A GPU implementation of the algorithm will probably alleviate this problem, and will be tackled as future work. Finally, other future work should aim at considering
 an appropriate surrogate-based approximation or different
initialization method and ad-hoc heuristics to keep the appropriate
exploration-exploitation balance with an affordable number of fitness evaluations
(actual simulations).

\section*{Acknowledgements}
J.-M.P. and A.L. acknowledge funding from the Spanish Ministry of Science
(TIN2017-83132-C2-2-R) and Universidad Polit{\'e}cnica de Madrid
(PINV-18-XEOGHQ-19-4QTEBP). M.T.K. and A.J. acknowledge funding from the European Research Council under the European Union's Seventh Framework Programme (FP7 2007–2013)/European Research Council Grant Agreement No. 306587. 

\bibliographystyle{elsarticle-num} 
\bibliography{neurite,thesis}






\par\noindent 
\parbox[t]{\linewidth}{
\noindent\parpic{\includegraphics[height=1.5in,width=1in,clip,keepaspectratio]{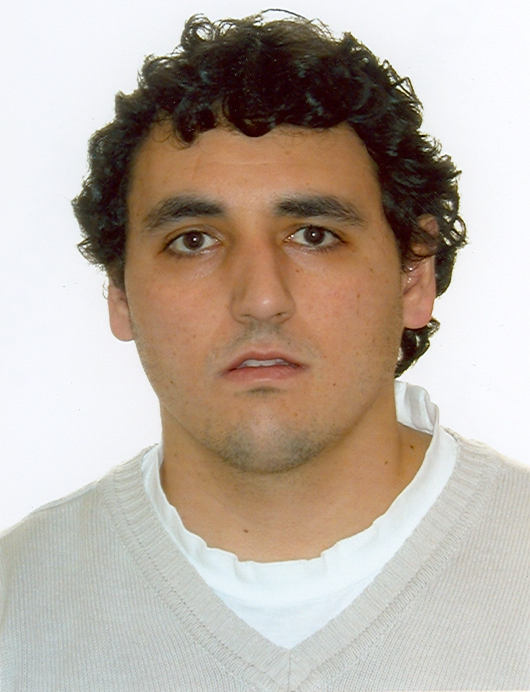}}
\noindent {\bf Antonio LaTorre}\
(PhD) is assistant profesor at UPM and subdirector of the CCS. His research interests focus on high-performance data analysis, modeling and optimization. He has an active research in applied problems in the domain of logistics, neurosciences and health. He has more than 14 years of research experience backed-up by his participation in 14 national and international projects, both with public and private funding, leading 3 of them. He has published more than 50 peer-reviewed contributions in international journals and conferences. He currently serves as vice-chair of the IEEE CIS Task Force on Large Scale Global Optimization.
}
\vspace{4\baselineskip}

\par\noindent 
\parbox[t]{\linewidth}{
\noindent\parpic{\includegraphics[height=1.5in,width=1in,clip,keepaspectratio]{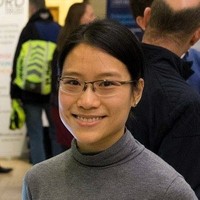}}
\noindent {\bf Man Ting Kwong}\
is a biomedical engineer currently working as a Post-doctoral Researcher at the University of Oxford. Her research interests are on the understanding and management of traumatic brain injuries. She obtained her MEng from the University of Sheffield. After working as a Research Assistant for two years at the University of Liverpool, she then moved on to completing her DPhil from the University of Oxford.
}
\vspace{4\baselineskip}

\par\noindent 
\parbox[t]{\linewidth}{
\noindent\parpic{\includegraphics[height=1.5in,width=1in,clip,keepaspectratio]{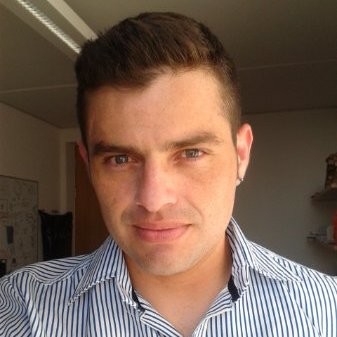}}
\noindent {\bf Juli{\'a}n A. Garc{\'i}a-Grajales}\
is a Senior Performance Simulation Engineer at Jaguar Formula E Team from 2016 in charge of full energy management, optimizations and race strategy. He worked as a Post-Doctoral Researcher at The University of Oxford focused on continuum modelling of axonal growth (2014-2016). He got his PhD on Computational Mechanical Modelling of Neurons, and his Master Engineering Science on Advanced Computing for Science and Engineering, both at the Technical University of Madrid (2010-2014). He got his degree as Aerospace Engineer with specialty on airplanes at the Technical University of Madrid (2004-2010).}
\vspace{4\baselineskip}

\par\noindent 
\parbox[t]{\linewidth}{
\noindent\parpic{\includegraphics[height=1.5in,width=1in,clip,keepaspectratio]{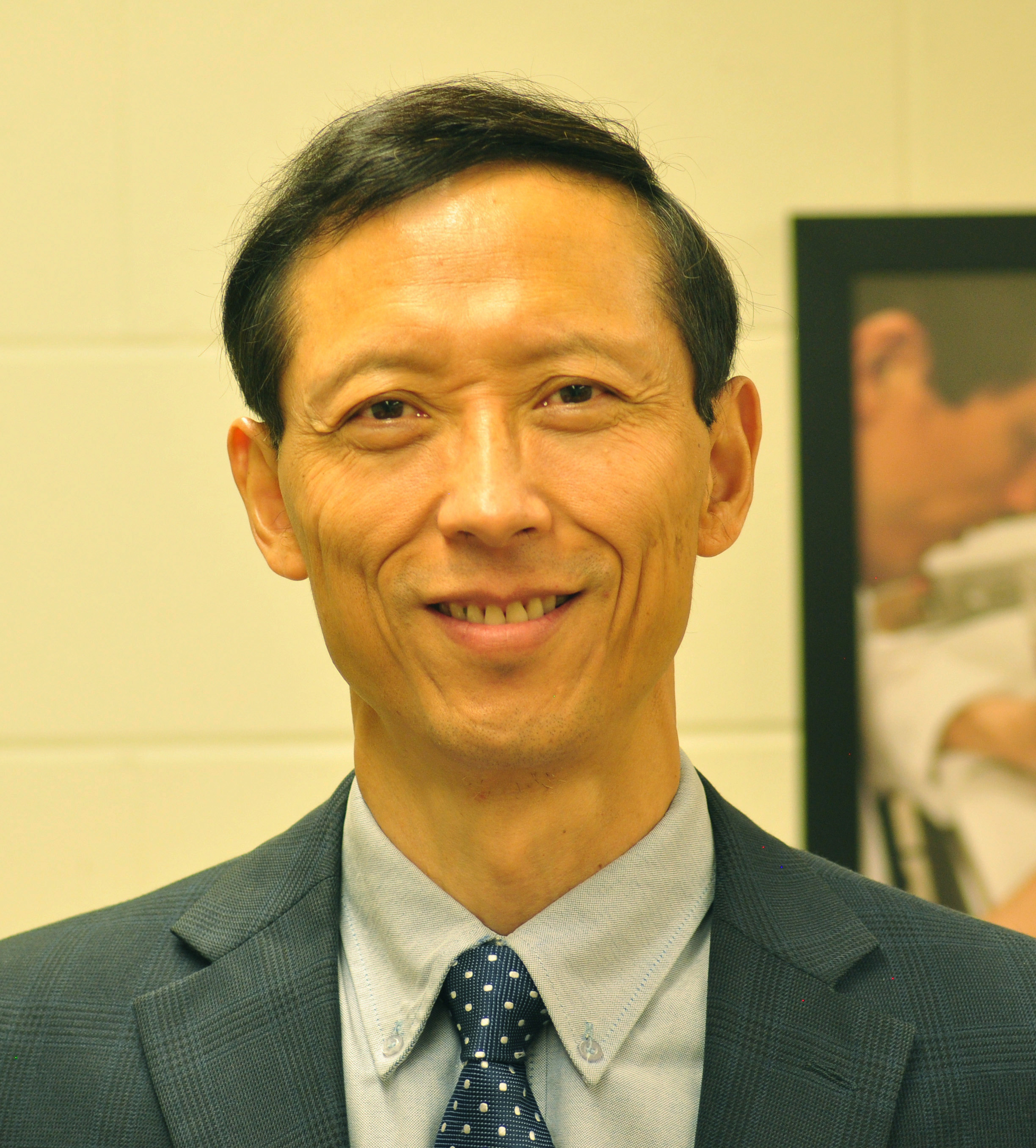}}
\noindent {\bf Riyi Shi}\
is currently a Professor of Neuroscience and Biomedical Engineering, specializing in uncovering the mechanisms of central nervous system trauma and diseases, and instituting new treatments. His contributions include originating the use of the double sucrose gap technique for recording action potential conduction, establishing polyethylene glycol (PEG) as a method of resealing neuronal membranes, identifying acrolein as a key pathological factor in neuronal trauma and degenerative diseases, and establishing a novel animal blast injury model. Before starting a research lab at Purdue University, he received his medical degree at the Shanghai Jiao-Tong University School of Medicine, Ph.D. from Purdue University, and postdoctoral training at the University of North Carolina at Chapel Hill.}
\vspace{4\baselineskip}

\par\noindent 
\parbox[t]{\linewidth}{
\noindent\parpic{\includegraphics[height=1.5in,width=1in,clip,keepaspectratio]{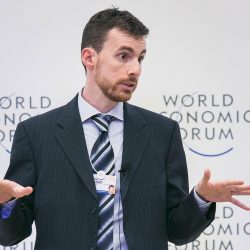}}
\noindent {\bf Antoine J{\'e}rusalem}\
graduated in 2004 with a double degree from the Ecole Nationale Supérieure de l’Aéronautique et de l’Espace with a Diplôme d’Ingénieur, and from the Massachusetts Institute of Technology with a M.Sc. in Aeronautics and Astronautics. In 2007, he obtained his Ph.D. in Computational Mechanics of Materials from MIT, where he stayed as a Postdoc for a year. He was the group leader of the Computational Mechanics of Materials Group in Madrid’s Advanced Studies Institute of Materials from 2008 to 2012, and is currently an Associate Professor in the Department of Engineering Science of the University of Oxford.}
\vspace{4\baselineskip}

\par\noindent 
\parbox[t]{\linewidth}{
\noindent\parpic{\includegraphics[height=1.5in,width=1in,clip,keepaspectratio]{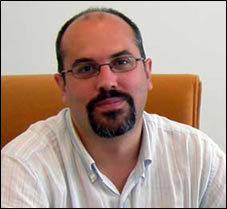}}
\noindent {\bf Jos\'{e} Mar\'{i}a Pe\~{n}a}\
(PhD) is Director of Lurtis Ltd. and Affiliated Scholar to the Computational Mechanics of Materials Group. He is also Associated Professor of the Universidad Politecnica de Madrid (UPM) and former Deputy Director of the Supercomputing and Visualization Center of Madrid (CeSViMa). He is member of the Intelligent Data Analysis (IDA) Council, associate and invited editor in several scientific journals and chair in conferences and workshops in the areas of data analysis, optimisation, simulation and distributed systems as well as in interdisciplinary conferences in bio/neuroinformatics. Dr. Peña has published more than 150 scientific journal and peer-reviewed conference papers.}
\vspace{4\baselineskip}

\end{document}